\documentclass{article} 
\usepackage{iclr2025_conference,times}

\usepackage{amsmath,amsfonts,bm}

\def\eqref#1{equation~\ref{#1}}

\def\1{\bm{1}}

\DeclareMathAlphabet{\mathsfit}{\encodingdefault}{\sfdefault}{m}{sl}
\SetMathAlphabet{\mathsfit}{bold}{\encodingdefault}{\sfdefault}{bx}{n}

\usepackage{hyperref}
\usepackage{url}

\definecolor{linkColor}{rgb}{0.18,0.39,0.62}

\definecolor{deepblue}{rgb}{0,0,0.5}
\definecolor{officeblue}{RGB}{0,102,204}
\definecolor{deepred}{rgb}{0.6,0,0}
\definecolor{deepgreen}{rgb}{0,0.5,0}
\definecolor{mybrickred}{RGB}{182,50,28}
\definecolor{nick_orange}{RGB}{255, 127, 80}
\definecolor{myred}{rgb}{0.992,0.9576,0.932}
\definecolor{mydred}{rgb}{0.992,0.915,0.892}
\definecolor{mypink}{rgb}{1,0.95,0.962}
\definecolor{myyellow}{rgb}{0.99,1,0.78}
\definecolor{myredd}{rgb}{0.992,0.9076,0.63}
\definecolor{mydredd}{rgb}{0.96,0.72,0.72}
\definecolor{mypinkd}{rgb}{0.98,0.75,0.952}

\usepackage{threeparttable}
\usepackage{multirow}
\usepackage{booktabs}
\usepackage{makecell}
\usepackage{xspace}

\usepackage{colortbl}

\usepackage{graphicx}
\usepackage{subfigure}
\usepackage{subcaption} 
\usepackage{wrapfig}
\usepackage{enumitem}
\setitemize{itemsep=10pt,topsep=0pt,parsep=0pt,partopsep=0pt}
\usepackage{pifont}

\def\eg{{\it{e.g.}}}

\def\ie{{\it{i.e.}}}
\def\etc{{\it{etc}}}

\newcommand{\fox}{\raisebox{-1.5pt}{\includegraphics[height=1.25em]{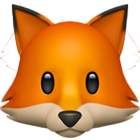}}\xspace}

\title{Unhackable Temporal Rewarding for Scalable Video MLLMs}

\author{En Yu$^1$\textsuperscript{,\P} \quad Kangheng Lin$^2$\textsuperscript{,\P} \quad Liang Zhao$^3$\textsuperscript{,\P} \quad Yana Wei$^4$ \quad Zining Zhu$^5$ \quad Haoran Wei$^3$ \\
\textbf{Jianjian Sun$^3$ \quad Zheng Ge$^3$  \quad Xiangyu Zhang$^3$ \quad Jingyu Wang$^2$$\footnotemark[2]$ \quad Wenbing Tao$^1$$\footnotemark[2]$} \\ 
$^1$Huazhong University of Science and Technology\\
$^2$Beijing University of Posts and Telecommunications \quad $^3$StepFun \\
$^4$Johns Hopkins University \quad $^5$University of Chinese Academy of Sciences \\ \\
 {\quad \quad \quad \quad \quad \quad \quad \quad \quad \quad \quad \quad \quad \fox \ \ Project Page: \href{https://Ahnsun.github.io/UTR/}{{\tt\text{Video-UTR}}}}
}

\iclrfinalcopy 
\begin{document}

\renewcommand{\thefootnote}{\fnsymbol{footnote}}
\footnotetext[2]{Corresponding authors, \textsuperscript{\P} Core contribution}
\renewcommand{\thefootnote}{\arabic{footnote}}

\maketitle
\vspace{-5.5mm}
\begin{abstract}
In the pursuit of superior video-processing MLLMs, we have encountered a perplexing paradox: the “\textit{anti-scaling law}”, where more data and larger models lead to worse performance. This study unmasks the culprit: \textit{“\textbf{temporal hacking}”}, a phenomenon where models shortcut by fixating on select frames, missing the full video narrative. In this work, we systematically establish a comprehensive theory of temporal hacking, defining it from a \textit{reinforcement learning} perspective, introducing the \textit{\textbf{T}emporal \textbf{P}erp\textbf{l}exity (\textbf{TPL})} score to assess this misalignment, and proposing the \textit{\textbf{U}nhackable \textbf{T}emporal \textbf{R}ewarding (\textbf{UTR})} framework to mitigate the temporal hacking. Both theoretically and empirically, TPL proves to be a reliable indicator of temporal modeling quality, correlating strongly with frame activation patterns. Extensive experiments reveal that UTR not only counters temporal hacking but significantly elevates video comprehension capabilities. This work not only advances video-AI systems but also illuminates the critical importance of aligning proxy rewards with true objectives in MLLM development.

 \end{abstract}

\section{Introduction}
\label{intro}

The pursuit of artificial intelligence that emulates human-like reasoning has increasingly emphasized the role of System 2 cognitive processes—deliberate~\citep{r1, o1}, structured~\citep{gpt4o}, and temporally-aware reasoning~\citep{merlin,fei2024video}—in advancing multimodal large language models (MLLMs). While early MLLMs like GPT-4V~\citep{gpt4v}, LLaVAs~\citep{llava, llava1p5} demonstrated remarkable capabilities in static image understanding, their application to video understanding remains constrained by the inherent complexity of spatiotemporal dynamics, long-range context dependencies, and multimodal alignment. This motivates researchers to develop powerful video MLLMs for the open-source community.

\begin{figure*}[t]
\centering
\includegraphics[width=1.0\linewidth]{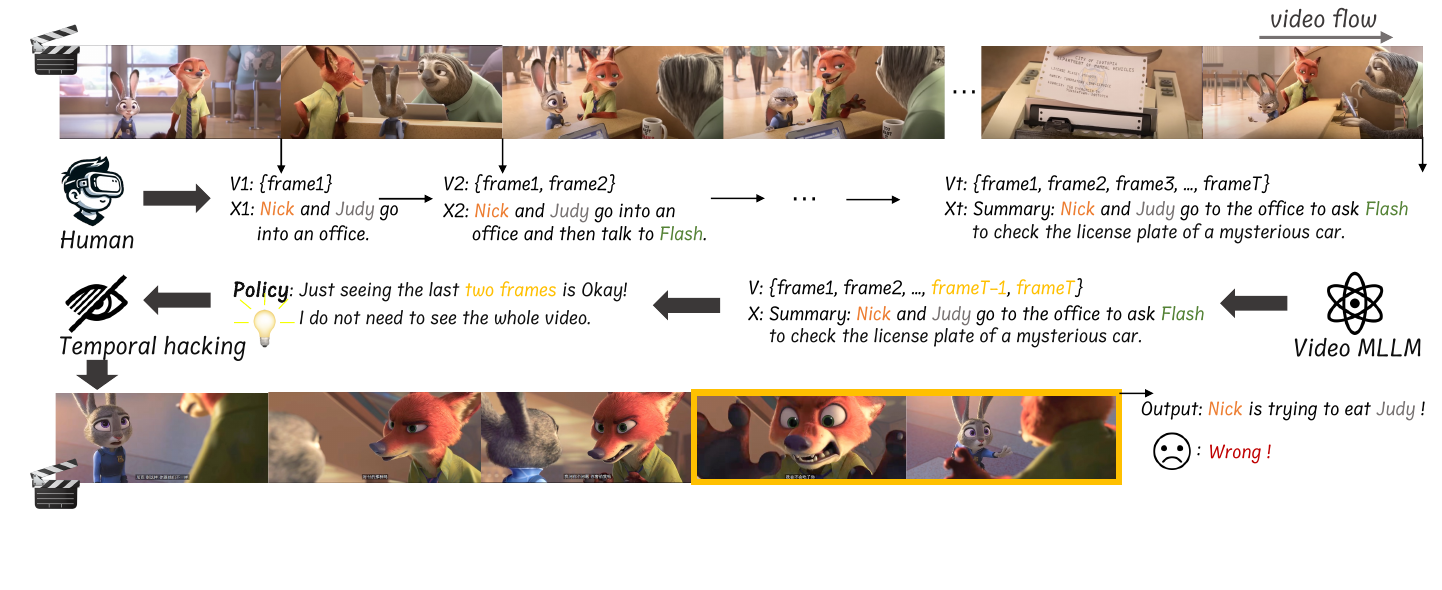}
\vspace{-12mm}
\caption{\textbf{Illustration of temporal hacking.} We select a scene from the \textit{Zootopia} to vividly illustrate the phenomenon of temporal hacking, where the fox is named \textbf{\textit{\textcolor{nick_orange}{Nick}}} and the rabbit is named \textbf{\textit{\textcolor{gray}{Judy}}}. Humans watch videos frame by frame, gradually building an understanding of the content, following a “flow” similar to a Markov process. In contrast, MLLMs process the entire video and its content at once, which can cause them to take shortcuts by focusing only on the most relevant frames.}
\label{fig:temporal_hacking}
\vspace{-6mm}
\end{figure*}

The dominant paradigm in video foundation model construction relies on contrastive~\citep{tong2022videomae, feichtenhofer2022masked, Internvideo2} or generative learning~\citep{videollama2, llavanext-video} from extensive video-text pair datasets. However, recent studies have unveiled a counterintuitive “\textit{anti-scaling law}” phenomenon~\citep{pllava}. Practically, increased data volume~\citep{wang2024tarsier} or model parameters~\citep{pllava} leads to performance degradation. Our analysis in Figure~\ref{fig:tp_vs_bmk} also shows that adding more training data decreases temporal modeling performance due to the dilution of high-quality samples. Further investigation reveals models often infer entire captions from a few key frames, typically just the initial (Figure~\ref{fig:attnmap}) or last one (Figure~\ref{fig:temporal_hacking}).
This suggests that current methodologies inadvertently promote a form of \textit{shortcut learning}. Critically, this issue resists resolution through mere data and parameter scaling. Such approaches may, in fact, exacerbate the problem. 

We propose to reframe this issue through the lens of \textit{reinforcement learning} (RL)~\citep{sutton2018reinforcement}. 
The generative modeling of MLLMs on video-text pairs can be formulated as a sequential decision-making process where the model’s policy aims to maximize the expected reward of generating highly relevant text conditioned on video frame context.
This formulation necessitates a critical examination: \textit{Does our proxy reward function (video-text or video-caption pair) adequately approximate the true reward (video-language alignment) we aim to optimize?}
Empirical evidence suggests a significant misalignment. We observe a manifestation of reward hacking~\citep{skalse2022defining} --- termed “\textbf{temporal hacking}” in the context of video LLMs. This predicament mirrors a boat in a racing game, furiously spinning in circles to collect “power-ups” while never advancing towards the finish line~\citep{boat}. 

Escaping the vortex of temporal reward hacking requires a shift in strategy, not merely increased effort. That is, \textbf{\textit{employing a more suitable proxy reward is key}} to overcoming this challenge. To this end, we first investigate the causes of temporal reward hacking and introduce a novel metric, \textit{\textbf{T}emporal \textbf{P}erp\textbf{l}exity (\textbf{TPL})} score, to quantify its severity. Experiments reveal a striking correlation between TP scores and models’ temporal modeling capabilities, with higher TPL scores consistently associated with the activation of more video frames. Our analysis further leads to the proposal of two key principles for designing an effective proxy reward function for video MLLMs: \textit{high frame information density} and \textit{high inter-frame information dynamics}. Guided by these two principles, we further propose an \textit{\textbf{U}nhackable \textbf{T}emporal \textbf{R}eward \textbf{(UTR)}}. UTR leverages \textit{spatiotemporal attributes} and \textit{bidirectional queries} to model video-language alignment. Comprehensive experiments validate that UTR, as an automated and scalable method, effectively achieves unhackable temporal modeling by guiding the model’s observational tendencies across all frames.

Our contributions are threefold:
\begin{itemize}
    \item We provide a novel RL perspective on the video MLLM unscaling phenomenon, systematically establishing \textit{“\textbf{temporal hacking}”} theory as its first comprehensive explanation.
    
    \item We design the \textit{\textbf{T}emporal \textbf{P}erp\textbf{l}exity (\textbf{TPL})} score, and through extensive experiments, TPL has demonstrated a high correlation with the true performance of the model, providing a reliable reference metric for mitigating temporal hacking.
    
    \item Through a series of theoretical and experimental analyses, we propose \textit{two principles} to guide the design of proxy rewards for video-language modeling and further propose \textit{\textbf{U}nhackable \textbf{T}emporal \textbf{R}ewarding (\textbf{UTR})}. Extensive experiments and analyses substantiate the effectiveness of UTR, offering crucial insights into video MLLM temporal modeling.
\end{itemize}

\section{Background \& Example Analysis}
\label{background&example}

\subsection{What is Temporal Hacking?}

\textbf{Reward hacking}~\citep{skalse2022defining, yuan2019novel}, also known as reward exploitation or reward gaming, refers to a phenomenon in reinforcement learning (RL) where an agent discovers a way to maximize its reward signal without actually achieving the intended goal of the task designer. Specifically, we first define a sequential decision problem $M = (S, A, P, R, \gamma)$, typically formalized as a Markov decision process (MDP), where $S$ is the state space, $A$ is the action space, $P: S \times A \times S \rightarrow [0, 1]$ is the transition probability function, $R: S \times A \times S \rightarrow \mathbb{R}$ is the reward function, and $\gamma \in [0, 1]$ is the discount factor. The goal of RL is to find a policy $\pi: S \rightarrow A$ that maximizes the expected cumulative discounted reward:
\vspace{-5mm}

\begin{equation}
\begin{aligned}
&\ J(\pi) = \mathbb{E}_{\tau \sim \pi} \left[ \sum\nolimits_{t=0}^{\infty} \gamma^t R(s_t, a_t) \right], \\
&\ \pi^{*} = \arg\max_{\pi} \mathbb{E}_{\tau \sim \pi} \left[\sum\nolimits_{t=0}^{\infty} \gamma^t R(s_t, a_t) \right], \\
\end{aligned}
\label{eq1}
\vspace{-3mm}
\end{equation}

where $\tau = (s_0, a_0, s_1, a_1, ...)$ is a trajectory generated by following policy $\pi$. $\pi^{*}$ is the optimal policy obtained under the current reward function. Reward hacking occurs when there exists a policy $\pi_h$ (generally $\pi_h = \pi^{*}$ ) such that:
\vspace{-5mm}

\begin{equation}
J(\pi_h) > J(\hat{\pi}), \ \ \text{\textit{however}}, \ \  K(\pi_h) \ll K(\hat{\pi}),
\label{eq2}
\vspace{-3mm}
\end{equation}

where $\hat{\pi}$ is the optimal policy for achieving the intended task, and $K$ denotes the true performance of the policy model in the intended task. In essence, reward hacking indicates an optimization misalignment, leading to policies that achieve high proxy rewards ($J(\pi_h)$) but fail to accomplish the true reward objectives ($K(\pi_h)$).

\textbf{From reward hacking to temporal hacking.} 
Autoregressive video-language modeling~\citep{li2023videochat, video-llama, llavanext-video}, aims to replicate human video comprehension. As illustrated in Figure~\ref{fig:temporal_hacking}, humans sequentially access each video frame, incrementally building an understanding by integrating all prior information~\citep{coltheart1980persistences}. Similarly, the model progressively generates tokens for each frame with the preceding video context conditioned. It is natural to represent this task as a sequential Markov decision process from an RL perspective.

Particularly, given a video frame sequence $ V = \{v_t\}_{t=1}^{T} $(where $T$ is the total number of frames) and a specific time step $t$, the sequence of preceding frames $V_{1:t}$ constitutes the state space, and the corresponding text token $x_t$ forms the action space. During training, the policy $\pi$ sequentially generates tokens $x_t$ conditioned on state $V_{1:t}$. The generated tokens’ quality and relevance to $V_{1:t}$ are evaluated by a reward function $R$, typically measured through the next token’s cross-entropy~\citep{gpt1, gpt2}. The objective can be formalized as:
\vspace{-5mm}

\begin{equation}
J(\pi) = \mathbb{E}_{\tau \sim \pi} \left[ \sum\nolimits_{t=1}^{T} \gamma^t R(V_{1:t}, x_t) \right].
\label{eq3}
\vspace{-3mm}
\end{equation}

By optimizing the policy model based on this objective function $J$, we obtain an optimal policy model $\pi^{*}$ under the current reward function. However, as shown in Figure~\ref{fig:temporal_hacking} and previous works~\citep{pllava,wang2024tarsier}, $\pi^{*}$ often fails to generate text that accurately aligns with video content and user instructions. Instead, the model may optimize the objective by \textit{accessing only a limited number of frames}, leading to shortcut learning. 
This issue, termed \textit{temporal hacking} in this paper, reflects the discrepancy between proxy and true objectives as described by Eq.~\ref{eq2}.

We provide an illustrative example in Figure~\ref{fig:temporal_hacking}, where it can be observed that the model, through temporal hacking, has identified a ``simpler" version of the true reward by focusing only on the last two frames of the video. This learned proxy reward can be highly dangerous in certain situations, leading to completely erroneous video understanding.

\subsection{What causes Temporal Hacking?}
\label{analysis_th}

In this section, we will analyze the causes of temporal hacking phenomenon in video-language modeling from both theoretical and experimental perspectives.

\textbf{Theoretical perspectives.} In reward hacking theory~\citep{skalse2022defining, yuan2019novel}, misalignment between proxy and true objectives ($J(\pi) \neq K(\pi)$) leads to shortcut learning. For video-language modeling, the true objective is to generate spatially and temporally comprehensive descriptions that align with human understanding of the video. However, in practice, the surrogate objective rewards consistency between model predictions and human-annotated captions~\citep{wang2024tarsier} or curated internet content~\citep{Bain21,wang2023internvid}. This discrepancy can result in suboptimal model behavior.

Ideally, as illustrated in Eq.~\ref{eq3}, trajectories $\tau = (V_{1:1}, x_1, ..., V_{1:t}, x_t, ...)$ propagates along every frame in the temporal sequence, implicitly necessitating a textual descriptions comprehensively describe each frame. However, due to frame redundancy and annotation costs, the text is often conditioned only on a subset of frames or aggregated information from multiple frames, especially in some \textit{static} or \textit{low-motion} scenarios. It is particularly challenging to provide a distinct description for each frame.
Consequently, the policy’s trajectory becomes $\tau = (V_{1:1}, x_1, ..., V_{k:t}, x_t, ...)$ where $V_{k:t}$ represents any frame set satisfying description $x_t$, and is a subset of $V_{1:t}$. The resultant surrogate objective can be expressed as:
\vspace{-5mm}

\begin{equation}
J(\pi) = \mathbb{E}_{\tau \sim \pi} \left[ \sum\nolimits_{t=1, 1 \leq k \leq t}^{T} \gamma^t R(V_{k:t}, x_t) \right].
\label{eq4}
\vspace{-3mm}
\end{equation}

As illustrated in Figure~\ref{fig:temporal_hacking}, optimizing such a proxy is insufficient and prone to deviate from the true objective of comprehensive video understanding. This reward hacking can be quantified by subtracting Eq.~\ref{eq3} from Eq.~\ref{eq4}, yielding $\Delta \mathcal{R}$:
\vspace{-5mm}

\begin{equation} 
\Delta \mathcal{R} = \sum\nolimits_{t=1, 1 \leq k \leq t}^{T} \gamma^t \left( R(V_{1:t}, x_t) - R(V_{k:t}, x_t) \right).
\label{eq5}
\vspace{-3mm}
\end{equation}

From Eq.~\ref{eq5}, it is evident that as $t$ increases, or as the average subset size $k$ increases (indicating that video descriptions can be condensed to fewer frames), the reward gap widens. This elucidates the observed “anti-scaling law” phenomenon in existing video-language models, where performance degrades as video length increases.

\begin{figure*}[t!]
    \centering
    \subfigure[]{
        \includegraphics[width=0.38\columnwidth]{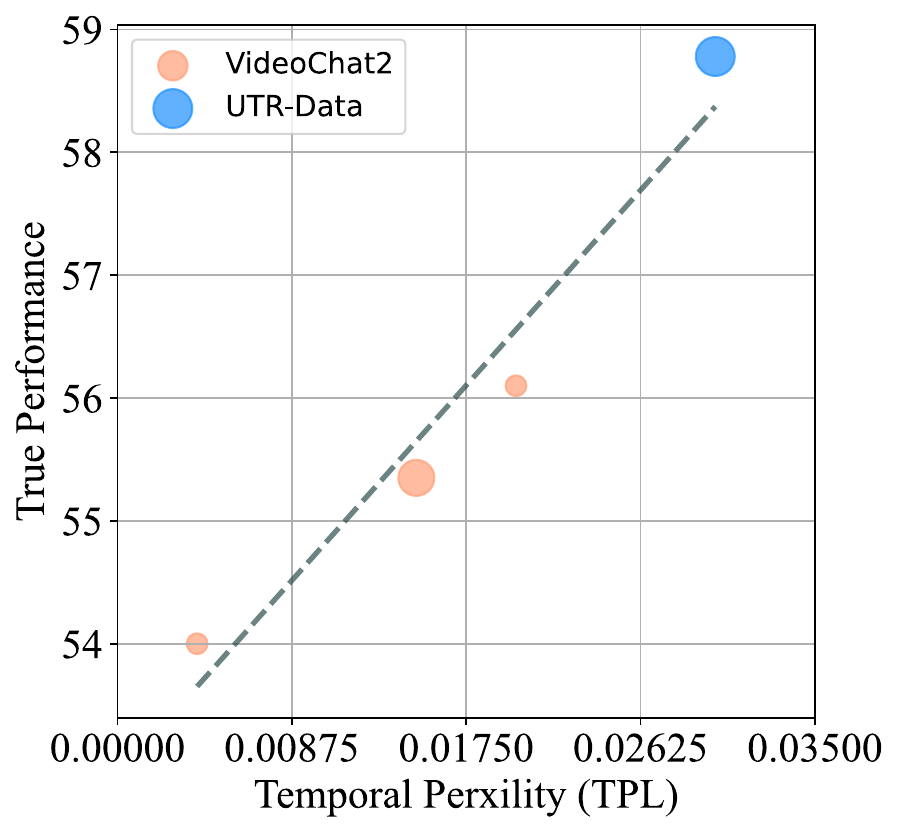}
        \label{fig:tp_vs_bmk}
    }
    \subfigure[]{
	\includegraphics[width=0.56\columnwidth]{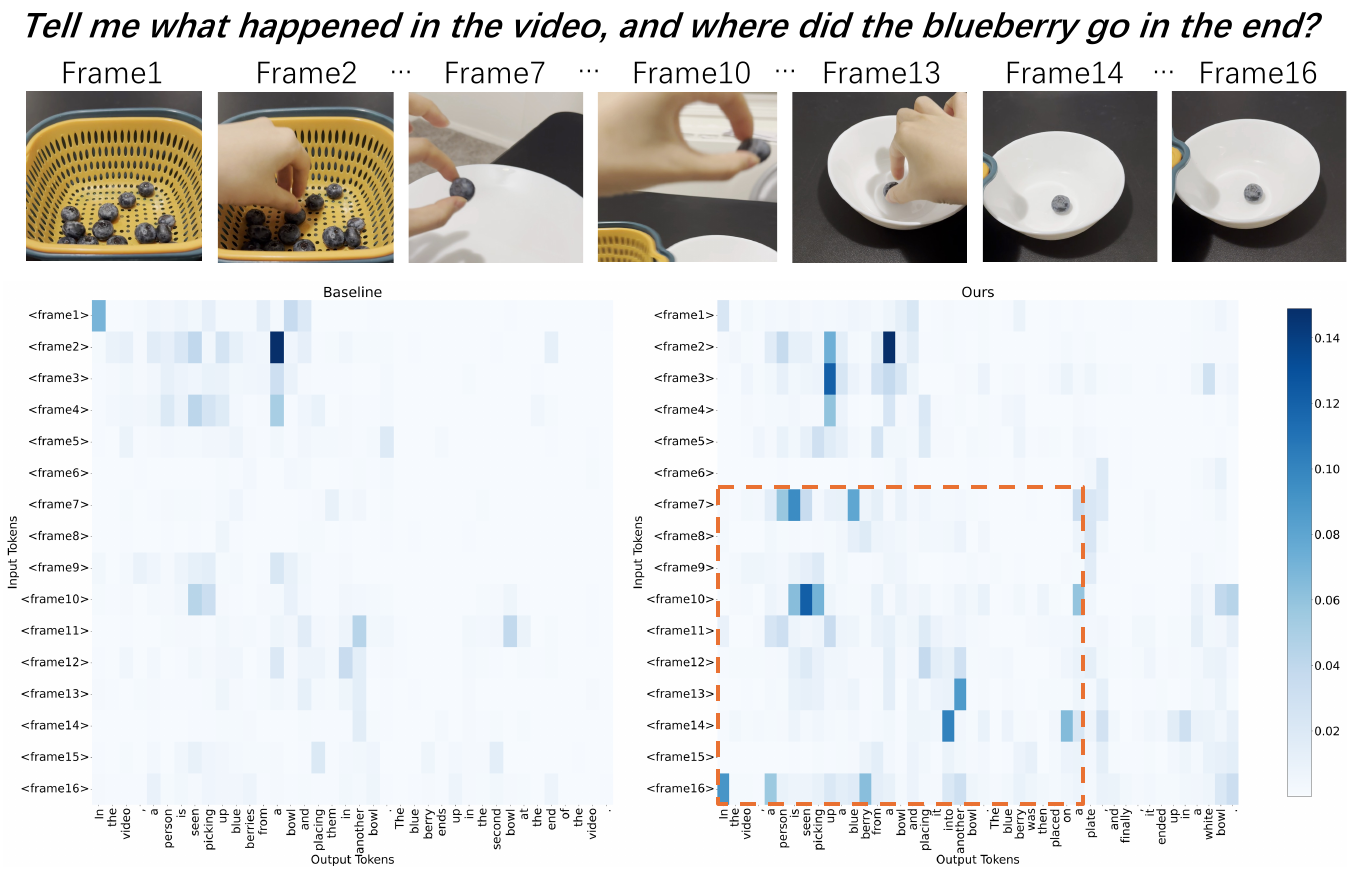}
        \label{fig:attnmap}
    }
    \vspace{-5mm}
        \caption{\textbf{Analysis of the temporal hacking.} \textbf{(a)} shows the relationship between temporal perplexity and true performance. The size of the radius of the circle represents the amount of data. \textbf{(b)} visualizes the attention map illustrating which specific frames the model’s output focuses on.}
        \vspace{-6mm}
        \label{fig:anaysis_tp}
\end{figure*}

\textbf{Experimental perspectives.} To shed light on reward hacking, we propose an extreme perspective to probe $\Delta \mathcal{R}$. We leverage perplexity~\citep{li2024superfiltering} $\mathcal{R}_{ppl}$ to model the cumulative reward between video context and its textual description. Higher similarity correlates with greater cumulative reward and lower model perplexity.
We simulate the true cumulative reward using a fully sampled video sequence as video context. To model an extreme case of proxy cumulative reward, we use a single, randomly sampled keyframe to represent the entire video context (\ie, $k = t$). This simulates a scenario where the model attempts to describe the whole video based on minimal information. The difference between these two rewards is defined in this paper as \textit{temporal perplexity (TPL, defined as $\mathcal{T}_{tpl}$)} or \textit{temporal hackability}. Formally,
\vspace{-5mm}

\begin{equation} 
\mathcal{T}_{tpl} = - \left(\mathcal{R}_{ppl}(V_{1:T}, x_T) - \mathcal{R}_{ppl}(V_{T:T}, x_T)\right).
\label{eq:tp} 
\vspace{-3mm}
\end{equation}

In practice, to avoid distributional shift, we utilize our own MLLM model, trained on the full set of video data, to calculate perplexity. We record the mean negative log-likelihood (NLL) loss across all text tokens for each sample (\ie,  the logarithm of perplexity) to represent $R_{\text{ppl}}$.

By combining Eq.~\ref{eq5} and Eq.~\ref{eq:tp}, we can intuitively infer that, under the same training setup, a lower TPL score indicates a larger $\Delta R$, which in turn leads to a more severe occurrence of temporal hacking. To prove this, we conduct two experiments as shown in Figure~\ref{fig:anaysis_tp} for in-depth analysis of the relation between TPL score and temporal hacking.


Specifically, we first fine-tuned models using subsets from VideoChat2~\citep{mvbench} data with varying $\mathcal{T}_{tpl}$ ranges and then mixed the data with different TPL. Intuitively, \textit{higher average TPL scores indicate a reduced likelihood of reward hacking, thereby leading to superior video comprehension performance.} Figure~\ref{fig:tp_vs_bmk} corroborates this, showing a significant correlation between video performance and TPL scores across multiple benchmarks, indicating that temporal perplexity effectively measures $\Delta \mathcal{R}$ and even reward hacking. Furthermore, we can also observe that when the TPL score is low, increasing the amount of data does not lead to performance gains, indicating the occurrence of the anti-scaling law phenomenon.

Then we delved deeper by analyzing attention maps of models on identical video-text pairs. Figure~\ref{fig:attnmap} illustrates that models trained on data with higher average-$\mathcal{T}_{tpl}$ activate more frames during inference on these well-described data. Conversely, models with lower-$\mathcal{T}_{tpl}$, due to severe reward hacking and inferior video modeling, activate fewer frames. These experiments demonstrate that our TPL score can effectively reflect the extent of temporal hacking, providing a reliable metric for exploring strategies to address this issue.

\section{Unhackable Temporal Rewarding}
\label{USTM}

\subsection{How to Mitigate Temporal Hacking?}

Section~\ref{background&example} introduces, defines, and analyzes the concept of \textit{temporal hacking}. A novel metric, \textit{temporal perplexity (TPL score)}, is proposed to assess whether the issue of temporal hacking arises in video-language modeling. At this point, the next important question arises: \textit{How can temporal hacking be mitigated or prevented?} Building upon the aforementioned analysis, we first propose two principles to guide the design of an \textbf{unhackable} reward in video-language temporal modeling:

\quad \textit{\textbf{Principle I\label{p1}: High frame information density.} 
The content of the video text should uniquely correspond to as many frames as possible.}

\quad \textit{\textbf{Principle II\label{p2}: High inter-frame information dynamics.} 
Descriptions for different frames should be coherent and reflect temporal variations and event progression.}

The Principle~\hyperref[p1]{I}, as delineated by Eq.~\ref{eq5}, aims to mitigate the $\Delta \mathcal{R}$ by reducing $k$ as discussed in Section~\ref{analysis_th}. This can be accomplished by ensuring each frame of the video is uniquely described. The Principle~\hyperref[p2]{II} emphasizes continuous dynamics, not only to further reduce $k$ and $\Delta \mathcal{R}$, but also to ensure the continuity of policy state transitions in Eq.~\ref{eq3}, thereby enhancing the model’s understanding of real-world physical laws.

Current temporal modeling approaches predominantly focus on maximizing the relevance and consistency of video information (Principle~\hyperref[p2]{II}). However, addressing Principle~\hyperref[p1]{I} remains challenging due to high frame rates and inter-frame redundancy, complicating textual descriptions of individual frames. Advanced techniques such as InternVID~\citep{wang2023internvid} and COSMO~\citep{wang2024cosmo} ameliorate information density to some extent through video interleave formats, yet they still struggle with the high information density of frames and fail to effectively model spatiotemporal dynamics, thus not fully addressing Principle~\hyperref[p2]{II}. Additionally, methods like COSA~\citep{chen2023cosa}, which concatenate image-text pairs to create video data, fail to establish spatiotemporal relationships between frames, entirely violating Principle~\hyperref[p2]{II}.

To simultaneously satisfy the two proposed principles, we further propose the 

\textit{\textbf{U}nhackable \textbf{T}emporal \textbf{R}ewarding \textbf{(UTR)}} to boostrap the video-language modeling.

\begin{figure*}[t]
\centering
\includegraphics[width=1.0\linewidth]{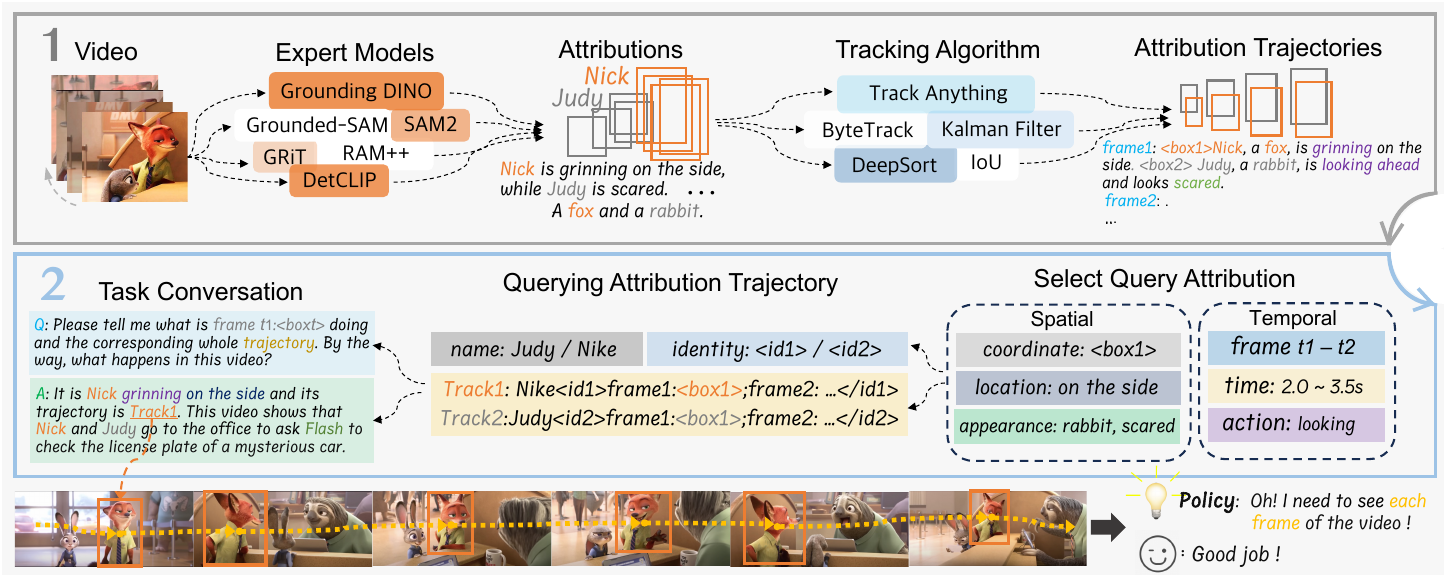}
\caption{\textbf{Overall pipeline of Unhackable Temporal Rewarding (UTR)}. UTR begins by using a mixture of expert models to extract unique spatiotemporal attributes and employs a tracking algorithm to construct multiple subject trajectories based on confidence levels (data modeling, top). It then performs bidirectional querying of temporal and spatial attributes to generate dialogue data (task modeling, bottom), thereby learning spatiotemporal dynamics.}
\label{fig:UTR}
\vspace{-5mm}
\end{figure*}

\subsection{Unhackable Temporal Rewarding}
\label{UTR}

As validated in Section~\ref{background&example}, suboptimal proxy rewards easily lead to temporal hacking in models. To address this, we propose a novel temporal rewarding method adhering to the aforementioned principles. Our approach, illustrated in Figure~\ref{fig:UTR}, extracts spatiotemporal attributes from video frames (row 1) and uniformly queries them (row 2) to model video-language alignment. This automated and scalable method achieves unhackable temporal modeling by guiding the model’s observational tendencies across all frames.

\textbf{Spatiotemporal attributes are key to representing unique video frame content.} Mitigating temporal hacking is challenging due to high frame rates and information redundancy in videos as mentioned before. We propose extracting \textit{spatiotemporal attributes} (e.g., trajectory, identity, action) to capture relatively independent information from each frame. This approach offers two advantages:

\begin{itemize}[leftmargin=2.5mm]
\setlength{\itemsep}{2pt}
\item \textit{Frame-to-frame variations in attributes, especially positional coordinates, enable modeling of frame-specific information, increasing information density (aligning with \textbf{Principle \hyperref[p1]{I})}}.

\item \textit{These attributes function as queries to link information across the video, facilitating learning of spatiotemporal dynamics (aligning with \textbf{Principle \hyperref[p2]{II})}}.

\end{itemize}

Specifically, given a video frame sequence $V = \{v_t\}_{t=1}^{T}$ with the same meaning in Eq.~\ref{eq3}, we extract the attribute information of subjects from each frame as follows:
\vspace{-5mm}

\begin{equation} 
X_{t} = \{x_t^{loc}, x_t^{app}, x_t^{act}\} = F(v_t),
\label{eq7}
\vspace{-3mm}
\end{equation}

where $x_t^{loc}, x_t^{app}, x_{t}^{act}$ indicate the location, appearance, and action information of subjects in frame $v_t$, respectively. Function $F$ extracts this information, using labeled data or specialized models such as GRiT~\citep{wu2022grit} and Grounding DINO~\citep{liu2023grounding}. We then organize this subject information into trajectories corresponding to each subject:
\vspace{-5mm}

\begin{equation} 
\{Y_{i}\}_{i=1}^{N} = \left\{\{y_{i, t}^{tr}, y_{i, t}^{id}, y_{i, t}^{act}\}_{t=1}^{T}\right\}_{i = 1}^{N}  = A(\{X_t\}_{t=1}^{T}),
\label{eq8} 
\vspace{-3mm}
\end{equation}

where $Y_{i}$ is the trajectory of subject $i$ and $N$ is the number of subjects in the video. To be specific, $y_{i, t}^{tr}, y_{i, t}^{id}, y_{i, t}^{act}$ indicate the trajectory, identity, and action information of subject $i$ in frame $v_t$, respectively. Function $A$ associates subjects across frames to form trajectories and identities, typically using tracking algorithms like ByteTrack~\citep{zhang2022bytetrack}.

\textbf{Bidirectional querying explicitly models spatiotemporal dynamics.} Previous methods~\citep{wang2023internvid,wang2024cosmo} modeled relatively dense information by interleaving text with selected frames, yet they neglected the critical spatiotemporal dynamics. Inspired by Merlin~\citep{merlin}, we propose a bidirectional querying mechanism that uses any temporal or spatial attribute to query global spatiotemporal attributes. This approach offers two benefits: 

\begin{itemize}[leftmargin=2.5mm]
\setlength{\itemsep}{2pt}

    \item \textit{Explicit modeling of spatiotemporal attributes forces the model to read each frame, aligning with \textbf{Principle \hyperref[p1]{I}.}}
    \item \textit{The arbitrariness of querying across time and space enhances the model’s understanding of spatiotemporal dynamics, and the stronger this arbitrariness, the deeper the understanding, aligning with \textbf{Principle \hyperref[p1]{II}.}}
    
\end{itemize}

Particularly, we randomly sample the information of one or more subjects as query attributes and select several frames as query frames. The model must predict the complete subject information based on the provided query data. Formally,
\vspace{-5mm}

\begin{equation}
P(Y|V,Y_{q}) \sim P\left(\{y_{s_i}\}_{i=1}^N|\{v_t\}_{t=1}^T, \{y_{s_i,t_j}\}_{i,j}\right),
\label{eq:bidirectional}
\vspace{-3mm}
\end{equation}

where $\{s_i\}_{i=1}^N \subseteq \{1,2,\ldots,N\}$ represents sampled subject identities, $\{t_j\}_{j=1}^M \subseteq \{1,2,\ldots,T\}$ indicates selected query frames, and $y_{s_i,t_j}$ denotes attribute information of selected subjects sampled from $Y$, which can be location, appearance, and action description. 

Notably, the random selection of query frames $\{t_j\}_{j=1}^M$ ensures the model utilizes query information from any part of the video—beginning, middle, or end—as cues to trace the entire trajectory. This approach not only compels the model to fully observe and comprehend the entire video, avoiding shortcuts like relying solely on initial or final frames, but also enhances its understanding of time-dependent physical laws. By necessitating the model to infer states across various temporal intervals, it implicitly learns concepts such as momentum, velocity, and acceleration, thereby strengthening its grasp of fundamental spatiotemporal dynamics.

\section{Empirical Result Details}
\label{exp}

\subsection{Experiment Settings}
\label{exp_setting}

\textbf{Datasets.} We primarily construct UTR-Data using several existing open-source video datasets, namely HowTo100M~\citep{miech2019howto100m}, MeViS~\citep{ding2023mevis}, and LaMOT~\citep{li2024lamot}. To extract subject attributions from each video frame, we use the region-to-text detector GRiT~\citep{wu2022grit}. Subsequently, we apply the ByteTrack~\citep{zhang2022bytetrack} tracking algorithm to construct attribution trajectories. Further details can be found in the Appendix~\ref{exp_detail}.

\textbf{Implementation Details.} To apply our UTR modeling strategy within the current video MLLM, we have developed a novel video MLLM, \ie, \textbf{\textit{Video-UTR}}. For the specific Video-UTR pipeline, we follow the general architecture in LLaVA-NeXT-Video~\citep{llavanext-video}, which consists of a vision encoder, SigLIP-L~\citep{zhai2023sigmoid}, a large language model, QWen-2~\citep{yang2024qwen2}, and a modality alignment projector, 2-layer GeLU-MLP. The training process consists of two stages. (1) \textit{Stage I}: Modality alignment, where only the projector is trained using the $558K$ LLaVA~\citep{llava} dataset. (2) \textit{Stage II}: Multi-task joint training, where the LLM is trained with various task datasets including video instruction-following data. Here, we mainly apply our \textbf{\textit{UTR}} in the \textit{Stage II}, which combines the constructed task data based on UTR with LLaVA-NeXT SFT data. Further details about the training settings can be found in the Appendix~\ref{exp_detail}.

\subsection{General Comprehension Evaluation}
\label{general_sota}

To showcase the generality and effectiveness of the proposed paradigm, we evaluated Video-UTR across various understanding benchmarks. Using the standard MLLM evaluation framework and the LLMs-Eval tool~\citep{zhang2024lmms}, we assessed major image and video understanding tasks. Results are shown in Tables~\ref{tab:general_video} and \ref{tab:general_image}. For video understanding, we focused on three general benchmarks: MVBench~\citep{mvbench}, TempCompass~\citep{tempcompass}, and VideoMME~\citep{videomme}, as well as four video QA benchmarks: MVSD-QA~\citep{mvsd}, MSRVTT-QA~\citep{msrvvt}, TGIF-QA~\citep{tgif}, and ActivityNet-QA~\citep{activitynet}. For image understanding, we reported scores from popular benchmarks like MM-Vet~\citep{mmvet}, MMBench~\citep{mmbench}, MMMU~\citep{mmmu}, MME~\citep{mme}, LLaVA-wild~\citep{llava}, SEED~\citep{seed}, AI2D~\citep{ai2d}, and RealWorldQA~\citep{grok}. For fairness, we used results from original papers.

\textbf{Video Understanding.} Table~\ref{tab:general_video} shows that Video-UTR outperforms other video MLLMs on most benchmarks, ranking first in 4 out of 7 tasks, highlighting its strong video understanding capabilities. Its high scores on MVBench ($58.78\%$), TempCompass ($59.67\%$), and VideoMME ($52.63\%$) demonstrate its ability to handle complex tasks like temporal reasoning, identifying differences, locating objects, tracking motion, and interpreting dynamic scenes. Additionally, its performance on four video QA benchmarks reflects exceptional understanding, particularly in managing temporally sensitive information. Remarkably, Video-UTR achieves these results using only about $1.1M$ video samples, a much smaller dataset compared to other models of similar performance, showcasing the efficiency and effectiveness of our UTR approach.

\textbf{Image Understanding.} Table~\ref{tab:general_image} shows that Video-UTR, despite being a video MLLM, delivers highly competitive performance compared to image-level MLLMs. For instance, on key benchmarks, Video-UTR matches or outperforms top image MLLMs like LLaVA-1.5~\citep{llava} ($39.6\%$ vs. $35.4\%$ on MM-Vet) and the stronger LLaVA-NeXT-Img~\citep{llavanext-video} ($76.6\%$ vs. $72.1\%$ on MMBench). It also performs well on hallucination benchmarks, achieving $88.9\%$ on POPE, and excels in image QA, with $63.7\%$ on RealWorldQA, showing its ability to avoid misidentification and misalignment with irrelevant image details. These results demonstrate that UTR not only helps video MLLMs overcome temporal hacking but also enhances their ability to analyze and understand images effectively.

\begin{table*}[t!]
\renewcommand{\arraystretch}{1.2}
  \centering
  \caption{\textbf{General Video Understanding Performance Comparsion} on 7 benchmarks, Video-UTR outperforms competitors in 4 out of 7 benchmarks and ranks second on the others, despite these competitors using larger training datasets or more parameters. Several benchmark names are abbreviated due to space limits. TempC: Tempcompass, ANet-QA: ActivityNet-QA. And Acc indicates Accuracy. The best results are \textbf{bold} and the second-best results are \underline{underlined}. $*$ indicates metrics reproduced by ourselves for evaluation cause original paper does not report.}
  \setlength{\tabcolsep}{1.05mm}{
  \resizebox{1.0\columnwidth}{!}{
    \begin{threeparttable} 
    \begin{tabular}{l c c | c  c c c c c c c c c c}
    \toprule
 	\multirow{2}{*}{Methods} & \multirow{2}{*}{LLM} & \text{Data} & \multirow{2}{*}{\textbf{MVBench}} & \multirow{2}{*}{\textbf{TempC}}  & \multirow{2}{*}{\textbf{VideoMME}} &  \multicolumn{2}{c} {\textbf{MSVD-QA}} &  \multicolumn{2}{c} {\textbf{MSRVVT-QA}} &  \multicolumn{2}{c} {\textbf{TGIF-QA}} &  \multicolumn{2}{c} {\textbf{ANet-QA}} \\
   \cmidrule(rl){7-8} \cmidrule(rl){9-10} \cmidrule(rl){11-12} \cmidrule(rl){13-14}

    & & \text{Scale} & & & & \text{Acc} & \text{Score}  & \text{Acc} & \text{Score} & \text{Acc} & \text{Score} & \text{Acc} & \text{Score}\\
    \midrule
    VideoChat~(\citeyear{li2023videochat}) & Vicuna-7B & $765K$ & 35.5 & $-$ & $-$ & 56.3 & 2.8 & 45.0 & 2.5 & 34.4 & 2.3 &  $-$ & 2.2 \\
    VideoChat2 (\citeyear{mvbench}) & Vicuna-7B & $1.9M$ & 51.1 & 38.5 &  $-$ & 70.0 & 3.9 & 54.1 & 3.3 & $-$ & $-$ & 49.1 & \underline{3.3} \\
    Video-ChatGPT (\citeyear{maaz2023video}) & Vicuna-7B& $765K$ & 32.7& 31.8&  $-$ & 51.6 & 2.5 & 29.6 & 1.8 & $-$ & $-$ & 12.4 & 1.1 \\
    Video-LLaVA (\citeyear{lin2023video}) & Vicuna-7B & $765K$ & 34.1 & 34.8& 39.9& 64.9 & 3.3 & 49.3 & 2.8 & 51.4 &3.0 & 35.2 & 2.7 \\
    VideoLLaMA2 (\citeyear{videollama2}) & LLaMA2-7B & $13.4M$ & 54.6 & $-$ & 46.6 & 70.9&\underline{3.8} & $-$ & $-$ & $-$ & $-$ & 50.2& \underline{3.3}\\
    PLLaVA (\citeyear{pllava}) & LLaMA2-7B & $1M$ & 46.6& $-$ &$-$ & \bf 76.6& \bf 4.1& \bf 62.0& \underline{3.5}& \bf 77.5& \bf 4.1& \underline{56.3}& 
 \bf3.5 \\
    LLaVA-NeXT-Video (\citeyear{llavanext-video}) & Qwen2-7B & $860K$ & 54.6 & $-$& 33.7 & 67.8& 3.5 & $-$& $-$& $-$& $-$& 53.5& 3.2\\
    LLaVA-OneVision(\citeyear{li2024llava}) & Qwen2-7B & $1.6M$ & \underline{56.7} & \underline{$59.0^{*}$}& \bf 58.2 & $65.3^{*}$ & \underline{$3.8^{*}$} & $43.3^{*}$ & $3.0^{*}$ & $52.8^{*}$ & $3.4^{*}$ & \bf$56.6^{*}$& \underline{$3.3^{*}$} \\
    \midrule
    \rowcolor{mydred}
    Video-UTR (\textbf{Ours}) & Qwen2-7B & $1.1M$ & \bf 58.8& \bf 59.7& \underline{52.6}&  \underline{73.5}& \bf 4.1& \underline{58.3}& \bf 3.6& \underline{56.4}& \underline{3.6}& 55.0 & 3.2\\
    \bottomrule
    \end{tabular}
    \end{threeparttable}}
}
  \label{tab:general_video}%
  \vspace{-3mm}
\end{table*}%
\begin{table*}[t!]
\renewcommand{\arraystretch}{1.2}
  \centering
  \caption{\textbf{General Image Understanding Performance Comparision} on 9 benchmarks, Video-UTR achieves performance comparable to, or even surpassing, that of pure image-level MLLMs. LLaVA\textsuperscript{W}: LLaVA in the wild. The best results are \textbf{bold} and the second-best results are \underline{underlined}.}
  \setlength{\tabcolsep}{1.05mm}{
  \resizebox{0.97\columnwidth}{!}{
    \begin{threeparttable} 
    \begin{tabular}{l c | c c c c c c c c c}
    \toprule

 	\multirow{2}{*}{Methods} & \multirow{2}{*}{LLM} & \multirow{2}{*}{\textbf{MM-Vet}} & \multirow{2}{*}{\textbf{MMBench}} & \multirow{2}{*}{\textbf{MMMU}} &  \multirow{2}{*}{\textbf{MME}}  & \multirow{2}{*}{\textbf{LLaVA}\textsuperscript{W}} & \multirow{2}{*}{\textbf{POPE}} & \multirow{2}{*}{\textbf{SEED}} & \multirow{2}{*}{\textbf{AI2D}} & \multirow{2}{*}{\textbf{RealWorldQA}} \\

    & & & & & & & & &\\
    \midrule
    \textit{Image-level MLLM} &&&&&&&&&\\
     InstructBLIP~(\citeyear{instructblip}) & Vicuna-7B & 33.1 & 36.0& 30.6   & 1137.1  & 59.8   &  86.1 & 53.4 & 40.6 & 36.9\\ 
     Qwen-VL-Chat~(\citeyear{bai2023qwen}) & Qwen-7B & \bf 47.3 &60.6 &  37.0    & 1467.8  &  67.7  &  74.9 &  58.2 & 63.0 & 49.3\\
      LLaVA-v1.5-7B~(\citeyear{llava1p5}) & Vicuna-7B & 30.5 & 64.3&  35.7   & 1510.7 & 61.8  & 86.1 & 58.6 & 55.5 & 54.8\\
     LLaVA-v1.5-13B & Vicuna-13B& 35.4 & 67.7 &  37.0   & 1531.3 & 66.1   & 88.4  &   61.6 & 61.1 & 55.3 \\
      ShareGPT4V~(\citeyear{chen2023sharegpt4v}) & Vicuna-7B & 37.6 & 68.8 & 37.2 & \underline{1567.4} & 72.6 & 86.6 & \underline{69.7} & 58.0 & 54.9 \\
     LLaVA-NeXT-Img~(\citeyear{llavanext-video}) & LLaMA3-8B & \underline{44.4} & \underline{72.1} & 41.7 & 1551.5 & 63.1 & 87.1 & $-$ & 71.6 & 60.0 \\ 
     \midrule
     \textit{Video-level MLLM} &&&&&&&&&\\
     LLaMA-VID~(\citeyear{llamavid}) & Vicuna-7B & $-$ & 66.6 & $-$  & 1521.4 & $-$ & 86.0  & 59.9 & $-$ & $-$ \\
     Video-LLaVA~(\citeyear{lin2023video}) & Vicuna-7B & 32.0 & 60.9 & $-$  & $-$ & \underline{73.1} & 84.4  & $-$ & $-$ & $-$ \\
    LLaVA-NeXT-Video~(\citeyear{llavanext-video}) & QWen2-7B & 42.9 & 74.5 & \underline{42.6}  & 1580.1 & \bf 75.9 & \underline{88.7}  & 74.6 & \underline{71.9} & \underline{60.1} \\
    \rowcolor{mydred}
    Video-UTR (\textbf{Ours}) & Qwen2-7B & 39.6 & \bf 76.6 & \bf 43.4  & \bf 1583.6 & 69.4 & \bf88.9  &  \bf 74.7 & \bf 72.1 & \bf63.7\\
    \bottomrule
    \end{tabular}

    \end{threeparttable}}
}
  \label{tab:general_image}%
\vspace{-6mm}
\end{table*}%

\subsection{Abalation Study about UTR}
\label{ablation}

\textbf{Effectiveness of each component of UTR.} In this ablation study, we evaluate the impact of removing the two key components of UTR: data modeling (UTR-Data) and task modeling (Bidirectional Querying) from Video-UTR. We focus on three major video and image understanding benchmarks. As shown in Table~\ref{tab:ablation}, removing UTR-Data and Bidirectional Querying leads to a significant drop in performance on video understanding tasks, emphasizing their importance in handling complex video reasoning tasks. Notably, removing UTR-Data causes a more consistent and pronounced decline across all benchmarks, including both image and video tasks. This underscores the critical role of \textit{data modeling} in UTR, as it directly aligns with the \textit{two principles} we proposed.

At the same time, to eliminate the potential influence of video data volume, we also add an equivalent amount of VideoChat2~\citep{mvbench} data. It can be observed that the additional video data did not result in further gains, which further underscores the importance of data modeling. Video data constructed in an improper manner will inevitably lead to temporal hacking, thus hindering the improvement of true video understanding performance.

\textbf{Ablation on the scalability of Video-UTR.} In Section~\ref{analysis_th}, we identify that the ``anti-scaling law" phenomenon observed in current video MLLMs is due to the issue of temporal hacking. To address this, we propose UTR as a mitigation strategy. In this experiment, we will demonstrate whether video MLLMs, with the integration of UTR, can exhibit scalability. As shown in Table~\ref{tab:ablation_scale_data} and Table~\ref{tab:ablation_scale_frame}, thanks to the incorporation of UTR, Video-UTR demonstrates a certain degree of scalability in the size of video data. Specifically, the larger the volume of video data, the better the model’s performance. Additionally, we observe that under the condition of unchanged video content, an increased number of video frames does not negatively impact the model’s performance. This scalability is advantageous for further exploring the better performance of Video-UTR in the future.

\begin{table*}[t!]
\renewcommand{\arraystretch}{1.2}
  \centering
  \caption{\textbf{Abalation study of Video-UTR} on both video and image understanding benchmarks.}
  \vspace{-2mm}
  \setlength{\tabcolsep}{2mm}{
    \resizebox{0.9\columnwidth}{!}{
    \begin{threeparttable} 
    \small
    \begin{tabular}{l c | c  c c | c c c }
    \toprule

	Ablation Setting & Data Scale &  \textbf{MVBench} & \textbf{TGIF-QA} & \textbf{ANet-QA}  & \textbf{MMVet} & \textbf{MMBench} & \textbf{POPE} \\
    \midrule

    \rowcolor{mydred}
     Video-UTR &    1.1M & \bf 58.78  & \bf 56.44  & \bf 55.00 & 39.59  & \bf 76.63  & 88.86     \\
\rowcolor{mydred!70}
     - Task Modeling& 1.0M & 58.45  & 56.11 & 54.21  & 37.33  & 76.37  & \bf 89.29\\
     \rowcolor{mydred!40}
     \ \ \ \ - Data Modeling & 780K & 54.63  & 54.74 & 54.15  & \bf 42.20  & 75.77  & 89.13 \\
     \ \ \ \ \ \ + More VideoChat2& 1.1M & 57.65  & 53.39  &  53.65 &  36.56 & 75.95  &  88.76 \\

    \bottomrule
    \end{tabular}

    \end{threeparttable}}
}
  \label{tab:ablation}%
\vspace{-2mm}
\end{table*}%

\begin{table}[t]
  \centering

\begin{minipage}[t]{0.48\textwidth}
\makeatletter\def\@captype{table}
\small
\centering
\caption{\textbf{Scalability of video data size.}}
\vspace{-2mm}
\resizebox{1.0\columnwidth}{!}{
\begin{tabular}{c|ccc}
\toprule
UTR-Data Size & \textbf{MVBench} & \textbf{TempCompass} & \textbf{VideoMME} \\
\midrule
0K & 54.63 & 58.88 &\bf 53.37 \\
180K &  58.45 & 58.47 & 52.30 \\
325K & \bf 58.78 & \bf 59.67 & 52.63 \\
\bottomrule
\end{tabular}
}
\label{tab:ablation_scale_data}

\end{minipage}
\begin{minipage}[t]{0.49\textwidth}
\makeatletter\def\@captype{table}
\small
\caption{\textbf{Scalability of frame length.}}
\vspace{-2mm}
\resizebox{1.0\columnwidth}{!}{
\centering
\begin{tabular}{c|ccc}
\toprule
Frame Length & \textbf{MVBench} & \textbf{TempCompass} & \textbf{VideoMME} \\
\midrule
8 & 50.93 & \bf 56.28 & 52.56 \\
24 & 50.08 & 56.14 & \bf 52.81 \\
32 & \bf 51.40 & 56.11 & 52.07 \\
\bottomrule
\end{tabular}
}
\label{tab:ablation_scale_frame}
\vspace{-3mm}


\end{minipage}
\vspace{-5mm}
\end{table}

\subsection{Spatial-temporal Understanding of Video-UTR}
\label{exp_st}

\begin{wrapfigure}{r}{0.56\textwidth}
    \vspace{-0.45cm}
    \centering
    \setlength\tabcolsep{5pt}
    \makeatletter\def\@captype{table}\makeatother
    \caption{\textbf{Zero-shot spatial-temporal understanding performance on MM-ID}~\citep{ji2024ida}.}\label{tab:ida_bench}
    \vspace{-0.25cm}
    \resizebox{\linewidth}{!}{
    \begin{tabular}{ccccc}
\toprule
\multicolumn{1}{c|}{Methods}      & \multicolumn{1}{l}{\textbf{Matching} $\uparrow$} & \multicolumn{1}{l}{\textbf{Location} $\uparrow$} & \multicolumn{1}{l}{\textbf{Q\&A} $\uparrow$} & \multicolumn{1}{l}{\textbf{Caption} $\uparrow$} \\ \midrule
\multicolumn{5}{l}{\textit{Open-source Models}}                                                                                                                                                     \\ \midrule
\multicolumn{1}{c|}{MMICL~(\citeyear{zhao2023mmicl})}               & $-$                                    & $-$                                     & 3.53                              & 3.18                                 \\
\multicolumn{1}{c|}{SEED~(\citeyear{ge2023making})}                & $-$                                     & $-$                                     & 3.19                              & 3.58                                 \\
\multicolumn{1}{c|}{QwenVL-Chat~(\citeyear{bai2023qwen})}         & $-$                                     & 0.504                                 & 3.63                              & 2.65                                 \\
\multicolumn{1}{c|}{InternLM-XComposer2~(\citeyear{dong2024internlm})} &                  $-$                      &            0.106                &       3.44                            &                 2.93              \\ \midrule
\multicolumn{5}{l}{\textit{Closed-source APIs}}                                                                                                                                                      \\ \midrule
\multicolumn{1}{c|}{QwenVL-Plus~(\citeyear{bai2023qwen})}         & 0.313                                 & 0.187                                 & 3.87                              & 3.79                                 \\
\multicolumn{1}{c|}{QwenVL-Max~(\citeyear{bai2023qwen})}          & 0.224                                 & 0.301                                 & 4.64                              & 4.23                                 \\
\multicolumn{1}{c|}{Gemini-pro~(\citeyear{gemini})}          & 0.687                                 & 0.081                                 & 4.97                              & 4.04                                 \\
\multicolumn{1}{c|}{GPT-4V~(\citeyear{gpt4v})}              & 0.627                                 & 0.244                                 & 4.77                              & 4.67                                 \\ \midrule
    \rowcolor{mydred}
\multicolumn{1}{c|}{Video-UTR (\textbf{Ours})}           & 0.277                                 & 0.328                                 & 4.36                              & 3.62              \\ \bottomrule
\end{tabular}
}
\vspace{0.2cm}

\centering
    \setlength\tabcolsep{5pt}
    \makeatletter\def\@captype{table}\makeatother
    \caption{\textbf{Zero-shot long-range video understanding performance on MMBench-Video}~\citep{fang2024mmbench}.}\label{tab:mmbench_video}
    \vspace{-0.25cm}
    \resizebox{\linewidth}{!}{
    \begin{tabular}{cccc}
\toprule
\multicolumn{1}{c|}{Methods}      & \multicolumn{1}{l}{\textbf{Overall} $\uparrow$} & \multicolumn{1}{l}{\textbf{Perception} $\uparrow$} & \multicolumn{1}{l}{\textbf{Reasoning} $\uparrow$} \\ \midrule                                                                      
\multicolumn{1}{c|}{Claude-3.5-Sonnet~(\citeyear{anthropic_claude_2024})}               & 1.35                                    & 1.4                                    & 1.04                                                      \\
\multicolumn{1}{c|}{VideoChat2-HD~(\citeyear{mvbench})}                & 1.23                                    & 0.44                                    & 1.23                                                             \\
\multicolumn{1}{c|}{PLLaVA-34B~(\citeyear{pllava})}         & 1.16                                    & 1                                 & 1.1                                                       \\
\multicolumn{1}{c|}{LLaVA-NeXT-Video-34B-HF~(\citeyear{llavanext-video})} &                  1.13                     &            0.58                &       1.03              \\ \midrule
    \rowcolor{mydred}
\multicolumn{1}{c|}{Video-UTR-7B (\textbf{Ours})}           & 1.35                                 & 1.38                                 & 1.24                            \\ \bottomrule
\end{tabular}
}

\vspace{-0.3cm}
\end{wrapfigure}

Spatial and temporal comprehension are equally important for multimodal video understanding. Here, we evaluate Video-UTR’s performance in these two areas using the latest benchmark, MM-ID, in a zero-shot setting. MM-ID tests a model’s ability to recognize identities across four increasingly complex levels, focusing on matching and locating objects with different identities across frames. As shown in Table~\ref{tab:ida_bench}, Video-UTR achieved highly competitive scores on both the matching and location sub-metrics without any MM-ID training data. Moreover, it outperformed methods using significantly larger datasets, further demonstrating the strength of the UTR approach. By leveraging spatiotemporal attribute modeling, UTR effectively enables the model to learn both spatial and temporal aspects.

In addition, we rigorously assess the performance of our Video-UTR on the recently introduced and exceptionally demanding long-range video understanding benchmark, MVBench-Video. As illustrated in Table~\ref{tab:mmbench_video}, our Video-UTR exhibits remarkable competitiveness on the performance of video perception and reasoning, surpassing a multitude of models with 34B parameters and even larger architectures, despite its compact 7B parameter size.

\subsection{In-depth Analysis about the TPL Score}
\label{exp_tp}

In Section~\ref{analysis_th}, we design a novel metric, temporal perplexity (TPL score), to measure the alignment degree between the proxy reward and the true reward. In this part, we aim to elucidate the correlation between TP scores and true rewards more intuitively from the perspective of video data quality. Specifically, we randomly select 100 video-text pairs from WebVid~\citep{Bain21} and calculate their temporal perplexity based on the definition in Eq.~\ref{eq:tp}. Then we pick two representative examples to illustrate the relationship between temporal perplexity (TPL) and the quality of video-text pairs.

\begin{figure*}[t]
\centering
\includegraphics[width=1.0\linewidth]{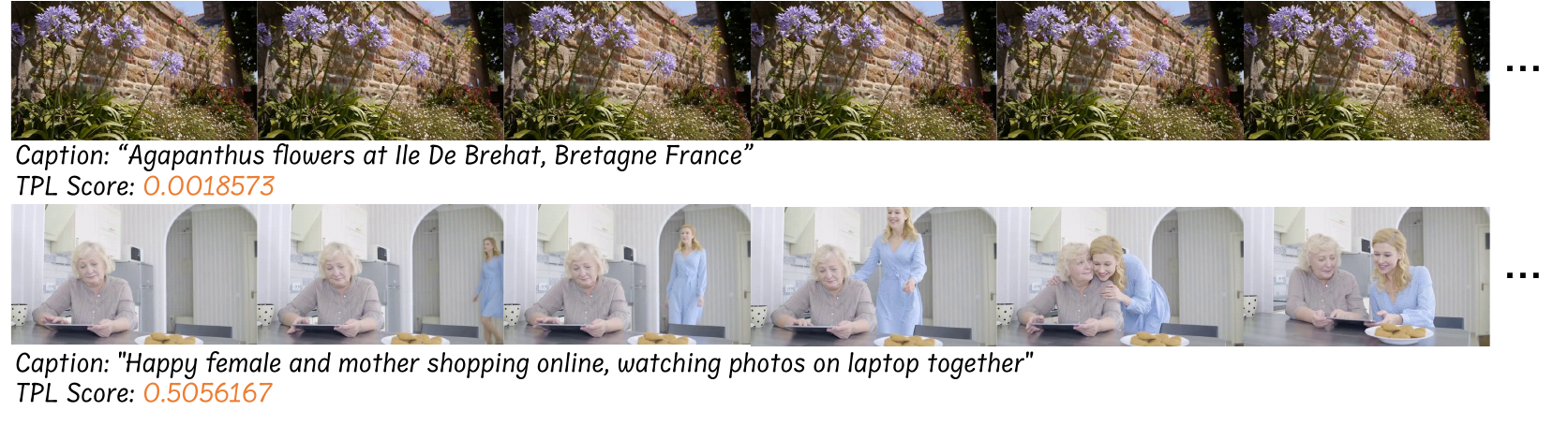}
\vspace{-8mm}
\caption{\textbf{Quantitative comparison} of the video-text pair with different temporal perplexity}
\label{fig:tpl_data}
\vspace{-6mm}
\end{figure*}

As shown in Figure~\ref{fig:tpl_data}, it can be observed that \textit{higher TPL score indicates a higher information density in the video or a more detailed description}. In this scenario, the model struggles to describe the entire video using just a single frame. Conversely, if the model’s performance based on a single frame is nearly as good as when using all frames, it either suggests that the video’s dynamics are negligible, making it almost like an image, or the textual description is so sparse that additional video information does not significantly improve modeling. The result aligns with our discussion in Section~\ref{analysis_th} and the analysis in Figure~\ref{fig:anaysis_tp}. This case study demonstrates that the TPL score can serve as a useful metric for filtering high-quality video-text pair data. Please refer to Appendix~\ref{add_exp} for more in-depth investigation. Furthermore, for a more comprehensive and in-depth analysis and discussion of the UTR methodology, please refer to Appendix~\ref{add_disscusion}.

\section{Related Work}
\label{related_work}

\textbf{Multimodal video foundation models.} Recently, vision-language models~\citep{llava, minigpt4, chatspot, wei2024small, wei2024vary} have demonstrated versatile visual understanding through visual instruction tuning~\citep{llava, minigpt4}. However, real-world video comprehension in multimodal models is still in its early stages. Due to the absence of powerful video encoders in the community, existing mainstream video MLLMs~\citep{llavanext-video, video-llama, videollama2} still rely on established images encoder, \ie, CLIP~\citep{clip}, to extract visual information frame by frame. Subsequently, they integrate temporal modeling techniques, \eg, Q-former~\citep{video-llama}, 3D Conv~\citep{videollama2}, and Pooling~\citep{llavanext-video, pllava}, to compress the visual tokens before feeding them with language tokens into LLMs. 

In addition to advancing the design of powerful temporal modules, recent works have increasingly acknowledged the pivotal role of \textit{video-language modeling} in video comprehension. Some works try to design filtering mechanisms~\citep{wang2024tarsier} to obtain high-quality video data with fine-grained description, while others aim to construct appropriate data structures~\citep{wang2023internvid, wang2024cosmo, chen2023cosa} and task formats~\citep{merlin} to enhance modeling performance. In this work, we systematically present how to design effective video-language modeling from a reinforcement learning perspective and propose guiding principles along with example frameworks.

\textbf{Reward hacking theory} was firstly introduced in the field of RL as a special case of Goodhart's Law~\citep{goodhart1984monetary, leike2018scalable}, and later explored in the context of AI alignment~\citep{leike2017ai}. ~\citep{rewardhacking} formalizes reward hacking by identifying types of reward mis-specifications that lead to it. Subsequent works~\citep{pan2022effects, laidlaw2024preventing} try to deal with reward hacking from different aspects. On the other hand, reward hacking is not exclusive to reinforcement learning, it also occurs in the optimization of pretrained visual generation models, where approaches often optimize towards a reward model by directly backpropagating gradients from a differentiable reward model~\citep{li2024reward, zhang2024large}. In this work, we transfer the concept of reward hacking to video-language modeling and establish a novel \textit{temporal hacking} theory to explain the shortcut learning in the existing video MLLM.

\section{Conclusion}
\label{conclusion}

In this work, we propose the theory of \textbf{\textit{temporal hacking}} from a reinforcement learning perspective to explain shortcut learning in video MLLMs. We introduce a novel metric, \textit{\textbf{T}emporal \textbf{P}erp\textbf{l}exity (\textbf{TPL})}, to quantify the severity of temporal hacking. Through extensive experiments, we use the TPL score to analyze the causes and features of temporal hacking, leading to the development of two guiding \textit{principles} for video-language modeling. We further propose \textit{\textbf{U}nhackable \textbf{V}ideo-Language \textbf{M}odeling (\textbf{UTR})} and build a powerful video MLLM, \ie, \textit{\textbf{Video-UTR}}. We hope this work offers a new perspective and insights to help the community build more robust video-AI systems.

\section*{Acknowledgements}

This work was supported by the National Natural Science Foundation of China under Grant 62176096.


\bibliography{iclr2025_conference}

\begin{thebibliography}{93}
\providecommand{\natexlab}[1]{#1}
\providecommand{\url}[1]{\texttt{#1}}
\expandafter\ifx\csname urlstyle\endcsname\relax
  \providecommand{\doi}[1]{doi: #1}\else
  \providecommand{\doi}{doi: \begingroup \urlstyle{rm}\Url}\fi

\bibitem[Anthropic(2024)]{anthropic_claude_2024}
Anthropic.
\newblock Claude 3.5 sonnet.
\newblock \url{https://www.anthropic.com/news/claude-3-5-sonnet}, 2024.

\bibitem[Bai et~al.(2023{\natexlab{a}})Bai, Bai, Chu, Cui, Dang, Deng, Fan, Ge, Han, Huang, et~al.]{qwen}
Jinze Bai, Shuai Bai, Yunfei Chu, Zeyu Cui, Kai Dang, Xiaodong Deng, Yang Fan, Wenbin Ge, Yu~Han, Fei Huang, et~al.
\newblock Qwen technical report.
\newblock \emph{arXiv preprint arXiv:2309.16609}, 2023{\natexlab{a}}.

\bibitem[Bai et~al.(2023{\natexlab{b}})Bai, Bai, Yang, Wang, Tan, Wang, Lin, Zhou, and Zhou]{bai2023qwen}
Jinze Bai, Shuai Bai, Shusheng Yang, Shijie Wang, Sinan Tan, Peng Wang, Junyang Lin, Chang Zhou, and Jingren Zhou.
\newblock Qwen-vl: A frontier large vision-language model with versatile abilities.
\newblock \emph{arXiv preprint arXiv:2308.12966}, 2023{\natexlab{b}}.

\bibitem[Bain et~al.(2021)Bain, Nagrani, Varol, and Zisserman]{Bain21}
Max Bain, Arsha Nagrani, G{\"u}l Varol, and Andrew Zisserman.
\newblock Frozen in time: A joint video and image encoder for end-to-end retrieval.
\newblock In \emph{IEEE International Conference on Computer Vision}, 2021.

\bibitem[Caba~Heilbron et~al.(2015)Caba~Heilbron, Escorcia, Ghanem, and Carlos~Niebles]{activitynet}
Fabian Caba~Heilbron, Victor Escorcia, Bernard Ghanem, and Juan Carlos~Niebles.
\newblock Activitynet: A large-scale video benchmark for human activity understanding.
\newblock In \emph{Proceedings of the ieee conference on computer vision and pattern recognition}, pp.\  961--970, 2015.

\bibitem[Chen et~al.(2023{\natexlab{a}})Chen, Li, Dong, Zhang, He, Wang, Zhao, and Lin]{chen2023sharegpt4v}
Lin Chen, Jisong Li, Xiaoyi Dong, Pan Zhang, Conghui He, Jiaqi Wang, Feng Zhao, and Dahua Lin.
\newblock Sharegpt4v: Improving large multi-modal models with better captions.
\newblock \emph{arXiv preprint arXiv:2311.12793}, 2023{\natexlab{a}}.

\bibitem[Chen et~al.(2023{\natexlab{b}})Chen, He, Li, Jin, Feng, and Liu]{chen2023cosa}
Sihan Chen, Xingjian He, Handong Li, Xiaojie Jin, Jiashi Feng, and Jing Liu.
\newblock Cosa: Concatenated sample pretrained vision-language foundation model.
\newblock \emph{arXiv preprint arXiv:2306.09085}, 2023{\natexlab{b}}.

\bibitem[Chen et~al.(2023{\natexlab{c}})Chen, Djolonga, Padlewski, Mustafa, Changpinyo, Wu, Ruiz, Goodman, Wang, Tay, et~al.]{chen2023pali}
Xi~Chen, Josip Djolonga, Piotr Padlewski, Basil Mustafa, Soravit Changpinyo, Jialin Wu, Carlos~Riquelme Ruiz, Sebastian Goodman, Xiao Wang, Yi~Tay, et~al.
\newblock Pali-x: On scaling up a multilingual vision and language model.
\newblock \emph{arXiv preprint arXiv:2305.18565}, 2023{\natexlab{c}}.

\bibitem[Chen et~al.(2024)Chen, Wu, Wang, Su, Chen, Xing, Zhong, Zhang, Zhu, Lu, et~al.]{chen2024internvl}
Zhe Chen, Jiannan Wu, Wenhai Wang, Weijie Su, Guo Chen, Sen Xing, Muyan Zhong, Qinglong Zhang, Xizhou Zhu, Lewei Lu, et~al.
\newblock Internvl: Scaling up vision foundation models and aligning for generic visual-linguistic tasks.
\newblock In \emph{Proceedings of the IEEE/CVF Conference on Computer Vision and Pattern Recognition}, pp.\  24185--24198, 2024.

\bibitem[Cheng et~al.(2024)Cheng, Leng, Zhang, Xin, Li, Chen, Zhu, Zhang, Luo, Zhao, et~al.]{videollama2}
Zesen Cheng, Sicong Leng, Hang Zhang, Yifei Xin, Xin Li, Guanzheng Chen, Yongxin Zhu, Wenqi Zhang, Ziyang Luo, Deli Zhao, et~al.
\newblock Videollama 2: Advancing spatial-temporal modeling and audio understanding in video-llms.
\newblock \emph{arXiv preprint arXiv:2406.07476}, 2024.

\bibitem[Coltheart(1980)]{coltheart1980persistences}
Max Coltheart.
\newblock The persistences of vision.
\newblock \emph{Philosophical Transactions of the Royal Society of London. B, Biological Sciences}, 290\penalty0 (1038):\penalty0 57--69, 1980.

\bibitem[Dai et~al.(2024)Dai, Li, Li, Tiong, Zhao, Wang, Li, Fung, and Hoi]{instructblip}
Wenliang Dai, Junnan Li, Dongxu Li, Anthony Meng~Huat Tiong, Junqi Zhao, Weisheng Wang, Boyang Li, Pascale~N Fung, and Steven Hoi.
\newblock Instructblip: Towards general-purpose vision-language models with instruction tuning.
\newblock \emph{Advances in Neural Information Processing Systems}, 36, 2024.

\bibitem[DeepMind(2018)]{rewardhacking}
DeepMind.
\newblock Specification gaming: The flaw in the reward, 2018.
\newblock URL \url{https://deepmind.google/discover/blog/specification-gaming-the-flip-side-of-ai-ingenuity/}.

\bibitem[DeepSeek-AI et~al.(2025)DeepSeek-AI, Guo, Yang, Zhang, Song, Zhang, Xu, Zhu, Ma, Wang, Bi, Zhang, Yu, Wu, Wu, Gou, Shao, Li, Gao, Liu, Xue, Wang, Wu, Feng, Lu, Zhao, Deng, Zhang, Ruan, Dai, Chen, Ji, Li, Lin, Dai, Luo, Hao, Chen, Li, Zhang, Bao, Xu, Wang, Ding, Xin, Gao, Qu, Li, Guo, Li, Wang, Chen, Yuan, Qiu, Li, Cai, Ni, Liang, Chen, Dong, Hu, Gao, Guan, Huang, Yu, Wang, Zhang, Zhao, Wang, Zhang, Xu, Xia, Zhang, Zhang, Tang, Li, Wang, Li, Tian, Huang, Zhang, Wang, Chen, Du, Ge, Zhang, Pan, Wang, Chen, Jin, Chen, Lu, Zhou, Chen, Ye, Wang, Yu, Zhou, Pan, Li, Zhou, Wu, Ye, Yun, Pei, Sun, Wang, Zeng, Zhao, Liu, Liang, Gao, Yu, Zhang, Xiao, An, Liu, Wang, Chen, Nie, Cheng, Liu, Xie, Liu, Yang, Li, Su, Lin, Li, Jin, Shen, Chen, Sun, Wang, Song, Zhou, Wang, Shan, Li, Wang, Wei, Zhang, Xu, Li, Zhao, Sun, Wang, Yu, Zhang, Shi, Xiong, He, Piao, Wang, Tan, Ma, Liu, Guo, Ou, Wang, Gong, Zou, He, Xiong, Luo, You, Liu, Zhou, Zhu, Xu, Huang, Li, Zheng, Zhu, Ma, Tang, Zha, Yan, Ren, Ren, Sha, Fu, Xu, Xie, Zhang,
  Hao, Ma, Yan, Wu, Gu, Zhu, Liu, Li, Xie, Song, Pan, Huang, Xu, Zhang, and Zhang]{r1}
DeepSeek-AI, Daya Guo, Dejian Yang, Haowei Zhang, Junxiao Song, Ruoyu Zhang, Runxin Xu, Qihao Zhu, Shirong Ma, Peiyi Wang, Xiao Bi, Xiaokang Zhang, Xingkai Yu, Yu~Wu, Z.~F. Wu, Zhibin Gou, Zhihong Shao, Zhuoshu Li, Ziyi Gao, Aixin Liu, Bing Xue, Bingxuan Wang, Bochao Wu, Bei Feng, Chengda Lu, Chenggang Zhao, Chengqi Deng, Chenyu Zhang, Chong Ruan, Damai Dai, Deli Chen, Dongjie Ji, Erhang Li, Fangyun Lin, Fucong Dai, Fuli Luo, Guangbo Hao, Guanting Chen, Guowei Li, H.~Zhang, Han Bao, Hanwei Xu, Haocheng Wang, Honghui Ding, Huajian Xin, Huazuo Gao, Hui Qu, Hui Li, Jianzhong Guo, Jiashi Li, Jiawei Wang, Jingchang Chen, Jingyang Yuan, Junjie Qiu, Junlong Li, J.~L. Cai, Jiaqi Ni, Jian Liang, Jin Chen, Kai Dong, Kai Hu, Kaige Gao, Kang Guan, Kexin Huang, Kuai Yu, Lean Wang, Lecong Zhang, Liang Zhao, Litong Wang, Liyue Zhang, Lei Xu, Leyi Xia, Mingchuan Zhang, Minghua Zhang, Minghui Tang, Meng Li, Miaojun Wang, Mingming Li, Ning Tian, Panpan Huang, Peng Zhang, Qiancheng Wang, Qinyu Chen, Qiushi Du, Ruiqi Ge, Ruisong
  Zhang, Ruizhe Pan, Runji Wang, R.~J. Chen, R.~L. Jin, Ruyi Chen, Shanghao Lu, Shangyan Zhou, Shanhuang Chen, Shengfeng Ye, Shiyu Wang, Shuiping Yu, Shunfeng Zhou, Shuting Pan, S.~S. Li, Shuang Zhou, Shaoqing Wu, Shengfeng Ye, Tao Yun, Tian Pei, Tianyu Sun, T.~Wang, Wangding Zeng, Wanjia Zhao, Wen Liu, Wenfeng Liang, Wenjun Gao, Wenqin Yu, Wentao Zhang, W.~L. Xiao, Wei An, Xiaodong Liu, Xiaohan Wang, Xiaokang Chen, Xiaotao Nie, Xin Cheng, Xin Liu, Xin Xie, Xingchao Liu, Xinyu Yang, Xinyuan Li, Xuecheng Su, Xuheng Lin, X.~Q. Li, Xiangyue Jin, Xiaojin Shen, Xiaosha Chen, Xiaowen Sun, Xiaoxiang Wang, Xinnan Song, Xinyi Zhou, Xianzu Wang, Xinxia Shan, Y.~K. Li, Y.~Q. Wang, Y.~X. Wei, Yang Zhang, Yanhong Xu, Yao Li, Yao Zhao, Yaofeng Sun, Yaohui Wang, Yi~Yu, Yichao Zhang, Yifan Shi, Yiliang Xiong, Ying He, Yishi Piao, Yisong Wang, Yixuan Tan, Yiyang Ma, Yiyuan Liu, Yongqiang Guo, Yuan Ou, Yuduan Wang, Yue Gong, Yuheng Zou, Yujia He, Yunfan Xiong, Yuxiang Luo, Yuxiang You, Yuxuan Liu, Yuyang Zhou, Y.~X. Zhu,
  Yanhong Xu, Yanping Huang, Yaohui Li, Yi~Zheng, Yuchen Zhu, Yunxian Ma, Ying Tang, Yukun Zha, Yuting Yan, Z.~Z. Ren, Zehui Ren, Zhangli Sha, Zhe Fu, Zhean Xu, Zhenda Xie, Zhengyan Zhang, Zhewen Hao, Zhicheng Ma, Zhigang Yan, Zhiyu Wu, Zihui Gu, Zijia Zhu, Zijun Liu, Zilin Li, Ziwei Xie, Ziyang Song, Zizheng Pan, Zhen Huang, Zhipeng Xu, Zhongyu Zhang, and Zhen Zhang.
\newblock Deepseek-r1: Incentivizing reasoning capability in llms via reinforcement learning, 2025.
\newblock URL \url{https://arxiv.org/abs/2501.12948}.

\bibitem[Ding et~al.(2023)Ding, Liu, He, Jiang, and Loy]{ding2023mevis}
Henghui Ding, Chang Liu, Shuting He, Xudong Jiang, and Chen~Change Loy.
\newblock Mevis: A large-scale benchmark for video segmentation with motion expressions.
\newblock In \emph{Proceedings of the IEEE/CVF International Conference on Computer Vision}, pp.\  2694--2703, 2023.

\bibitem[Dong et~al.(2024{\natexlab{a}})Dong, Han, Peng, Qi, Ge, Yang, Zhao, Sun, Zhou, Wei, Kong, Zhang, Ma, and Yi]{dreamllm}
Runpei Dong, Chunrui Han, Yuang Peng, Zekun Qi, Zheng Ge, Jinrong Yang, Liang Zhao, Jianjian Sun, Hongyu Zhou, Haoran Wei, Xiangwen Kong, Xiangyu Zhang, Kaisheng Ma, and Li~Yi.
\newblock Dream{LLM}: Synergistic multimodal comprehension and creation.
\newblock In \emph{The Twelfth International Conference on Learning Representations}, 2024{\natexlab{a}}.

\bibitem[Dong et~al.(2024{\natexlab{b}})Dong, Zhang, Zang, Cao, Wang, Ouyang, Wei, Zhang, Duan, Cao, et~al.]{dong2024internlm}
Xiaoyi Dong, Pan Zhang, Yuhang Zang, Yuhang Cao, Bin Wang, Linke Ouyang, Xilin Wei, Songyang Zhang, Haodong Duan, Maosong Cao, et~al.
\newblock Internlm-xcomposer2: Mastering free-form text-image composition and comprehension in vision-language large model.
\newblock \emph{arXiv preprint arXiv:2401.16420}, 2024{\natexlab{b}}.

\bibitem[Fang et~al.(2024)Fang, Mao, Duan, Zhao, Li, Lin, and Chen]{fang2024mmbench}
Xinyu Fang, Kangrui Mao, Haodong Duan, Xiangyu Zhao, Yining Li, Dahua Lin, and Kai Chen.
\newblock Mmbench-video: A long-form multi-shot benchmark for holistic video understanding.
\newblock \emph{arXiv preprint arXiv:2406.14515}, 2024.

\bibitem[Fei et~al.(2024)Fei, Wu, Ji, Zhang, Zhang, Lee, and Hsu]{fei2024video}
Hao Fei, Shengqiong Wu, Wei Ji, Hanwang Zhang, Meishan Zhang, Mong-Li Lee, and Wynne Hsu.
\newblock Video-of-thought: Step-by-step video reasoning from perception to cognition.
\newblock In \emph{Forty-first International Conference on Machine Learning}, 2024.

\bibitem[Feichtenhofer et~al.(2022)Feichtenhofer, Li, He, et~al.]{feichtenhofer2022masked}
Christoph Feichtenhofer, Yanghao Li, Kaiming He, et~al.
\newblock Masked autoencoders as spatiotemporal learners.
\newblock \emph{Advances in neural information processing systems}, 35:\penalty0 35946--35958, 2022.

\bibitem[Fu et~al.(2023)Fu, Chen, Shen, Qin, Zhang, Lin, Qiu, Lin, Yang, Zheng, Li, Sun, and Ji]{mme}
Chaoyou Fu, Peixian Chen, Yunhang Shen, Yulei Qin, Mengdan Zhang, Xu~Lin, Zhenyu Qiu, Wei Lin, Jinrui Yang, Xiawu Zheng, Ke~Li, Xing Sun, and Rongrong Ji.
\newblock Mme: A comprehensive evaluation benchmark for multimodal large language models.
\newblock \emph{arXiv preprint arXiv:2306.13394}, 2023.

\bibitem[Fu et~al.(2024)Fu, Dai, Luo, Li, Ren, Zhang, Wang, Zhou, Shen, Zhang, et~al.]{videomme}
Chaoyou Fu, Yuhan Dai, Yondong Luo, Lei Li, Shuhuai Ren, Renrui Zhang, Zihan Wang, Chenyu Zhou, Yunhang Shen, Mengdan Zhang, et~al.
\newblock Video-mme: The first-ever comprehensive evaluation benchmark of multi-modal llms in video analysis.
\newblock \emph{arXiv preprint arXiv:2405.21075}, 2024.

\bibitem[Ge et~al.(2023{\natexlab{a}})Ge, Zhao, Zeng, Ge, Li, Wang, and Shan]{ge2023making}
Yuying Ge, Sijie Zhao, Ziyun Zeng, Yixiao Ge, Chen Li, Xintao Wang, and Ying Shan.
\newblock Making llama see and draw with seed tokenizer.
\newblock \emph{arXiv preprint arXiv:2310.01218}, 2023{\natexlab{a}}.

\bibitem[Ge et~al.(2023{\natexlab{b}})Ge, Zhao, Zeng, Ge, Li, Wang, and Shan]{seed}
Yuying Ge, Sijie Zhao, Ziyun Zeng, Yixiao Ge, Chen Li, Xintao Wang, and Ying Shan.
\newblock Making llama see and draw with seed tokenizer.
\newblock \emph{arXiv preprint arXiv:2310.01218}, 2023{\natexlab{b}}.

\bibitem[Goodhart(1984)]{goodhart1984monetary}
Charles Albert~Eric Goodhart.
\newblock Monetary theory and practice: The uk experience.
\newblock \emph{(No Title)}, 1984.

\bibitem[GPT-4o(2024)]{gpt4o}
GPT-4o.
\newblock Hello gpt-4o, 2024.
\newblock URL \url{https://openai.com/index/hello-gpt-4o/}.

\bibitem[Gupta et~al.(2019)Gupta, Dollar, and Girshick]{gupta2019lvis}
Agrim Gupta, Piotr Dollar, and Ross Girshick.
\newblock Lvis: A dataset for large vocabulary instance segmentation.
\newblock In \emph{Proceedings of the IEEE/CVF conference on computer vision and pattern recognition}, pp.\  5356--5364, 2019.

\bibitem[Jack \& Dario(2016)Jack and Dario]{boat}
Clark Jack and Amodei Dario.
\newblock Faulty reward functions in the wild, 2016.
\newblock URL \url{https://openai.com/index/faulty-reward-functions/}.
\newblock December 21, 2016.

\bibitem[Jang et~al.(2017)Jang, Song, Yu, Kim, and Kim]{tgif}
Yunseok Jang, Yale Song, Youngjae Yu, Youngjin Kim, and Gunhee Kim.
\newblock Tgif-qa: Toward spatio-temporal reasoning in visual question answering.
\newblock In \emph{Proceedings of the IEEE conference on computer vision and pattern recognition}, pp.\  2758--2766, 2017.

\bibitem[Ji et~al.(2024)Ji, Zhang, Wu, Sun, Chen, Xiao, Yang, Yang, and Luo]{ji2024ida}
Yatai Ji, Shilong Zhang, Jie Wu, Peize Sun, Weifeng Chen, Xuefeng Xiao, Sidi Yang, Yujiu Yang, and Ping Luo.
\newblock Ida-vlm: Towards movie understanding via id-aware large vision-language model.
\newblock \emph{arXiv preprint arXiv:2407.07577}, 2024.

\bibitem[Kembhavi et~al.(2016)Kembhavi, Salvato, Kolve, Seo, Hajishirzi, and Farhadi]{ai2d}
Aniruddha Kembhavi, Mike Salvato, Eric Kolve, Minjoon Seo, Hannaneh Hajishirzi, and Ali Farhadi.
\newblock A diagram is worth a dozen images.
\newblock In \emph{Computer Vision--ECCV 2016: 14th European Conference, Amsterdam, The Netherlands, October 11--14, 2016, Proceedings, Part IV 14}, pp.\  235--251. Springer, 2016.

\bibitem[Krishna et~al.(2017)Krishna, Zhu, Groth, Johnson, Hata, Kravitz, Chen, Kalantidis, Li, Shamma, et~al.]{krishna2017visual}
Ranjay Krishna, Yuke Zhu, Oliver Groth, Justin Johnson, Kenji Hata, Joshua Kravitz, Stephanie Chen, Yannis Kalantidis, Li-Jia Li, David~A Shamma, et~al.
\newblock Visual genome: Connecting language and vision using crowdsourced dense image annotations.
\newblock \emph{International journal of computer vision}, 123:\penalty0 32--73, 2017.

\bibitem[Laidlaw et~al.(2024)Laidlaw, Singhal, and Dragan]{laidlaw2024preventing}
Cassidy Laidlaw, Shivam Singhal, and Anca Dragan.
\newblock Preventing reward hacking with occupancy measure regularization.
\newblock \emph{arXiv preprint arXiv:2403.03185}, 2024.

\bibitem[Leike et~al.(2017)Leike, Martic, Krakovna, Ortega, Everitt, Lefrancq, Orseau, and Legg]{leike2017ai}
Jan Leike, Miljan Martic, Victoria Krakovna, Pedro~A Ortega, Tom Everitt, Andrew Lefrancq, Laurent Orseau, and Shane Legg.
\newblock Ai safety gridworlds.
\newblock \emph{arXiv preprint arXiv:1711.09883}, 2017.

\bibitem[Leike et~al.(2018)Leike, Krueger, Everitt, Martic, Maini, and Legg]{leike2018scalable}
Jan Leike, David Krueger, Tom Everitt, Miljan Martic, Vishal Maini, and Shane Legg.
\newblock Scalable agent alignment via reward modeling: a research direction.
\newblock \emph{arXiv preprint arXiv:1811.07871}, 2018.

\bibitem[Li et~al.(2024{\natexlab{a}})Li, Zhang, Guo, Zhang, Li, Zhang, Zhang, Li, Liu, and Li]{li2024llava}
Bo~Li, Yuanhan Zhang, Dong Guo, Renrui Zhang, Feng Li, Hao Zhang, Kaichen Zhang, Yanwei Li, Ziwei Liu, and Chunyuan Li.
\newblock Llava-onevision: Easy visual task transfer.
\newblock \emph{arXiv preprint arXiv:2408.03326}, 2024{\natexlab{a}}.

\bibitem[Li et~al.(2024{\natexlab{b}})Li, Feng, Chen, and Wang]{li2024reward}
Jiachen Li, Weixi Feng, Wenhu Chen, and William~Yang Wang.
\newblock Reward guided latent consistency distillation.
\newblock \emph{arXiv preprint arXiv:2403.11027}, 2024{\natexlab{b}}.

\bibitem[Li et~al.(2023{\natexlab{a}})Li, He, Wang, Li, Wang, Luo, Wang, Wang, and Qiao]{li2023videochat}
KunChang Li, Yinan He, Yi~Wang, Yizhuo Li, Wenhai Wang, Ping Luo, Yali Wang, Limin Wang, and Yu~Qiao.
\newblock Videochat: Chat-centric video understanding.
\newblock \emph{arXiv preprint arXiv:2305.06355}, 2023{\natexlab{a}}.

\bibitem[Li et~al.(2024{\natexlab{c}})Li, Wang, He, Li, Wang, Liu, Wang, Xu, Chen, Luo, et~al.]{mvbench}
Kunchang Li, Yali Wang, Yinan He, Yizhuo Li, Yi~Wang, Yi~Liu, Zun Wang, Jilan Xu, Guo Chen, Ping Luo, et~al.
\newblock Mvbench: A comprehensive multi-modal video understanding benchmark.
\newblock In \emph{Proceedings of the IEEE/CVF Conference on Computer Vision and Pattern Recognition}, pp.\  22195--22206, 2024{\natexlab{c}}.

\bibitem[Li et~al.(2024{\natexlab{d}})Li, Zhang, He, Li, Zhao, Wang, Cheng, and Zhou]{li2024superfiltering}
Ming Li, Yong Zhang, Shwai He, Zhitao Li, Hongyu Zhao, Jianzong Wang, Ning Cheng, and Tianyi Zhou.
\newblock Superfiltering: Weak-to-strong data filtering for fast instruction-tuning.
\newblock \emph{arXiv preprint arXiv:2402.00530}, 2024{\natexlab{d}}.

\bibitem[Li et~al.(2023{\natexlab{b}})Li, Wang, and Jia]{llamavid}
Yanwei Li, Chengyao Wang, and Jiaya Jia.
\newblock Llama-vid: An image is worth 2 tokens in large language models.
\newblock \emph{arXiv preprint arXiv:2311.17043}, 2023{\natexlab{b}}.

\bibitem[Li et~al.(2024{\natexlab{e}})Li, Liu, Liu, Fan, and Zhang]{li2024lamot}
Yunhao Li, Xiaoqiong Liu, Luke Liu, Heng Fan, and Libo Zhang.
\newblock Lamot: Language-guided multi-object tracking.
\newblock \emph{arXiv preprint arXiv:2406.08324}, 2024{\natexlab{e}}.

\bibitem[Lin et~al.(2023)Lin, Zhu, Ye, Ning, Jin, and Yuan]{lin2023video}
Bin Lin, Bin Zhu, Yang Ye, Munan Ning, Peng Jin, and Li~Yuan.
\newblock Video-llava: Learning united visual representation by alignment before projection.
\newblock \emph{arXiv preprint arXiv:2311.10122}, 2023.

\bibitem[Lin et~al.(2014)Lin, Maire, Belongie, Hays, Perona, Ramanan, Doll{\'a}r, and Zitnick]{coco}
Tsung-Yi Lin, Michael Maire, Serge Belongie, James Hays, Pietro Perona, Deva Ramanan, Piotr Doll{\'a}r, and C~Lawrence Zitnick.
\newblock Microsoft coco: Common objects in context.
\newblock In \emph{Computer Vision--ECCV 2014: 13th European Conference, Zurich, Switzerland, September 6-12, 2014, Proceedings, Part V 13}, pp.\  740--755. Springer, 2014.

\bibitem[Liu et~al.(2024{\natexlab{a}})Liu, Li, Li, and Lee]{llava1p5}
Haotian Liu, Chunyuan Li, Yuheng Li, and Yong~Jae Lee.
\newblock Improved baselines with visual instruction tuning.
\newblock In \emph{Proceedings of the IEEE/CVF Conference on Computer Vision and Pattern Recognition}, pp.\  26296--26306, 2024{\natexlab{a}}.

\bibitem[Liu et~al.(2024{\natexlab{b}})Liu, Li, Wu, and Lee]{llava}
Haotian Liu, Chunyuan Li, Qingyang Wu, and Yong~Jae Lee.
\newblock Visual instruction tuning.
\newblock \emph{Advances in neural information processing systems}, 36, 2024{\natexlab{b}}.

\bibitem[Liu et~al.(2023{\natexlab{a}})Liu, Zeng, Ren, Li, Zhang, Yang, Li, Yang, Su, Zhu, et~al.]{liu2023grounding}
Shilong Liu, Zhaoyang Zeng, Tianhe Ren, Feng Li, Hao Zhang, Jie Yang, Chunyuan Li, Jianwei Yang, Hang Su, Jun Zhu, et~al.
\newblock Grounding dino: Marrying dino with grounded pre-training for open-set object detection.
\newblock \emph{arXiv preprint arXiv:2303.05499}, 2023{\natexlab{a}}.

\bibitem[Liu et~al.(2023{\natexlab{b}})Liu, Duan, Zhang, Li, Zhang, Zhao, Yuan, Wang, He, Liu, et~al.]{mmbench}
Yuan Liu, Haodong Duan, Yuanhan Zhang, Bo~Li, Songyang Zhang, Wangbo Zhao, Yike Yuan, Jiaqi Wang, Conghui He, Ziwei Liu, et~al.
\newblock Mmbench: Is your multi-modal model an all-around player?
\newblock \emph{arXiv preprint arXiv:2307.06281}, 2023{\natexlab{b}}.

\bibitem[Liu et~al.(2024{\natexlab{c}})Liu, Li, Liu, Wang, Ren, Li, Chen, Sun, and Hou]{tempcompass}
Yuanxin Liu, Shicheng Li, Yi~Liu, Yuxiang Wang, Shuhuai Ren, Lei Li, Sishuo Chen, Xu~Sun, and Lu~Hou.
\newblock Tempcompass: Do video llms really understand videos?
\newblock \emph{arXiv preprint arXiv:2403.00476}, 2024{\natexlab{c}}.

\bibitem[Maaz et~al.(2023)Maaz, Rasheed, Khan, and Khan]{maaz2023video}
Muhammad Maaz, Hanoona Rasheed, Salman Khan, and Fahad~Shahbaz Khan.
\newblock Video-chatgpt: Towards detailed video understanding via large vision and language models.
\newblock \emph{arXiv preprint arXiv:2306.05424}, 2023.

\bibitem[Meta(2024)]{llama3.2}
Meta.
\newblock Llama3.2-vision, 2024.
\newblock URL \url{https://www.llama.com/}.

\bibitem[Miech et~al.(2019)Miech, Zhukov, Alayrac, Tapaswi, Laptev, and Sivic]{miech2019howto100m}
Antoine Miech, Dimitri Zhukov, Jean-Baptiste Alayrac, Makarand Tapaswi, Ivan Laptev, and Josef Sivic.
\newblock Howto100m: Learning a text-video embedding by watching hundred million narrated video clips.
\newblock In \emph{Proceedings of the IEEE/CVF international conference on computer vision}, pp.\  2630--2640, 2019.

\bibitem[OpenAI(2023)]{gpt4v}
OpenAI.
\newblock Gpt-4v(ision) system card.
\newblock \url{https://cdn.openai.com/papers/GPTV_System_Card.pdf}, 2023.

\bibitem[OpenAI(2024)]{o1}
OpenAI.
\newblock Learning to reason with llms, September 2024.
\newblock URL \url{https://openai.com/index/learning-to-reason-with-llms/}.

\bibitem[Pan et~al.(2022)Pan, Bhatia, and Steinhardt]{pan2022effects}
Alexander Pan, Kush Bhatia, and Jacob Steinhardt.
\newblock The effects of reward misspecification: Mapping and mitigating misaligned models.
\newblock \emph{arXiv preprint arXiv:2201.03544}, 2022.

\bibitem[Peng et~al.(2024)Peng, Cui, Tang, Qi, Dong, Bai, Han, Ge, Zhang, and Xia]{peng2024dreambench++}
Yuang Peng, Yuxin Cui, Haomiao Tang, Zekun Qi, Runpei Dong, Jing Bai, Chunrui Han, Zheng Ge, Xiangyu Zhang, and Shu-Tao Xia.
\newblock Dreambench++: A human-aligned benchmark for personalized image generation.
\newblock \emph{arXiv preprint arXiv:2406.16855}, 2024.

\bibitem[Radford et~al.(2018)Radford, Narasimhan, Salimans, Sutskever, et~al.]{gpt1}
Alec Radford, Karthik Narasimhan, Tim Salimans, Ilya Sutskever, et~al.
\newblock Improving language understanding by generative pre-training.
\newblock \emph{article}, 2018.

\bibitem[Radford et~al.(2019)Radford, Wu, Child, Luan, Amodei, Sutskever, et~al.]{gpt2}
Alec Radford, Jeffrey Wu, Rewon Child, David Luan, Dario Amodei, Ilya Sutskever, et~al.
\newblock Language models are unsupervised multitask learners.
\newblock \emph{OpenAI blog}, 1\penalty0 (8):\penalty0 9, 2019.

\bibitem[Radford et~al.(2021)Radford, Kim, Hallacy, Ramesh, Goh, Agarwal, Sastry, Askell, Mishkin, Clark, et~al.]{clip}
Alec Radford, Jong~Wook Kim, Chris Hallacy, Aditya Ramesh, Gabriel Goh, Sandhini Agarwal, Girish Sastry, Amanda Askell, Pamela Mishkin, Jack Clark, et~al.
\newblock Learning transferable visual models from natural language supervision.
\newblock In \emph{International conference on machine learning}, pp.\  8748--8763. PMLR, 2021.

\bibitem[Rafailov et~al.(2024)Rafailov, Sharma, Mitchell, Manning, Ermon, and Finn]{rafailov2024direct}
Rafael Rafailov, Archit Sharma, Eric Mitchell, Christopher~D Manning, Stefano Ermon, and Chelsea Finn.
\newblock Direct preference optimization: Your language model is secretly a reward model.
\newblock \emph{Advances in Neural Information Processing Systems}, 36, 2024.

\bibitem[Schulman et~al.(2017)Schulman, Wolski, Dhariwal, Radford, and Klimov]{schulman2017proximal}
John Schulman, Filip Wolski, Prafulla Dhariwal, Alec Radford, and Oleg Klimov.
\newblock Proximal policy optimization algorithms.
\newblock \emph{arXiv preprint arXiv:1707.06347}, 2017.

\bibitem[Skalse et~al.(2022)Skalse, Howe, Krasheninnikov, and Krueger]{skalse2022defining}
Joar Skalse, Nikolaus Howe, Dmitrii Krasheninnikov, and David Krueger.
\newblock Defining and characterizing reward gaming.
\newblock \emph{Advances in Neural Information Processing Systems}, 35:\penalty0 9460--9471, 2022.

\bibitem[Sutton \& Barto(2018)Sutton and Barto]{sutton2018reinforcement}
Richard~S. Sutton and Andrew~G. Barto.
\newblock \emph{{Reinforcement learning: An introduction}}.
\newblock MIT press, 2018.

\bibitem[Team et~al.(2023)Team, Anil, Borgeaud, Wu, Alayrac, Yu, Soricut, Schalkwyk, Dai, Hauth, et~al.]{gemini}
Gemini Team, Rohan Anil, Sebastian Borgeaud, Yonghui Wu, Jean-Baptiste Alayrac, Jiahui Yu, Radu Soricut, Johan Schalkwyk, Andrew~M Dai, Anja Hauth, et~al.
\newblock Gemini: a family of highly capable multimodal models.
\newblock \emph{arXiv preprint arXiv:2312.11805}, 2023.

\bibitem[Tong et~al.(2022)Tong, Song, Wang, and Wang]{tong2022videomae}
Zhan Tong, Yibing Song, Jue Wang, and Limin Wang.
\newblock Videomae: Masked autoencoders are data-efficient learners for self-supervised video pre-training.
\newblock \emph{Advances in neural information processing systems}, 35:\penalty0 10078--10093, 2022.

\bibitem[Wang et~al.(2024{\natexlab{a}})Wang, Li, Lin, Wang, Lin, Yang, Wang, and Shou]{wang2024cosmo}
Alex~Jinpeng Wang, Linjie Li, Kevin~Qinghong Lin, Jianfeng Wang, Kevin Lin, Zhengyuan Yang, Lijuan Wang, and Mike~Zheng Shou.
\newblock Cosmo: Contrastive streamlined multimodal model with interleaved pre-training.
\newblock \emph{arXiv preprint arXiv:2401.00849}, 2024{\natexlab{a}}.

\bibitem[Wang et~al.(2024{\natexlab{b}})Wang, Yuan, and Zhang]{wang2024tarsier}
Jiawei Wang, Liping Yuan, and Yuchen Zhang.
\newblock Tarsier: Recipes for training and evaluating large video description models.
\newblock \emph{arXiv preprint arXiv:2407.00634}, 2024{\natexlab{b}}.

\bibitem[Wang et~al.(2023)Wang, He, Li, Li, Yu, Ma, Li, Chen, Chen, Wang, et~al.]{wang2023internvid}
Yi~Wang, Yinan He, Yizhuo Li, Kunchang Li, Jiashuo Yu, Xin Ma, Xinhao Li, Guo Chen, Xinyuan Chen, Yaohui Wang, et~al.
\newblock Internvid: A large-scale video-text dataset for multimodal understanding and generation.
\newblock \emph{arXiv preprint arXiv:2307.06942}, 2023.

\bibitem[Wang et~al.(2024{\natexlab{c}})Wang, Li, Li, Yu, He, Chen, Pei, Zheng, Xu, Wang, et~al.]{Internvideo2}
Yi~Wang, Kunchang Li, Xinhao Li, Jiashuo Yu, Yinan He, Guo Chen, Baoqi Pei, Rongkun Zheng, Jilan Xu, Zun Wang, et~al.
\newblock Internvideo2: Scaling video foundation models for multimodal video understanding.
\newblock \emph{arXiv preprint arXiv:2403.15377}, 2024{\natexlab{c}}.

\bibitem[Wei et~al.(2024{\natexlab{a}})Wei, Kong, Chen, Zhao, Ge, Yang, Sun, Han, and Zhang]{wei2024vary}
Haoran Wei, Lingyu Kong, Jinyue Chen, Liang Zhao, Zheng Ge, Jinrong Yang, Jianjian Sun, Chunrui Han, and Xiangyu Zhang.
\newblock Vary: Scaling up the vision vocabulary for large vision-language model.
\newblock In \emph{European Conference on Computer Vision}, pp.\  408--424. Springer, 2024{\natexlab{a}}.

\bibitem[Wei et~al.(2024{\natexlab{b}})Wei, Kong, Chen, Zhao, Ge, Yu, Sun, Han, and Zhang]{wei2024small}
Haoran Wei, Lingyu Kong, Jinyue Chen, Liang Zhao, Zheng Ge, En~Yu, Jianjian Sun, Chunrui Han, and Xiangyu Zhang.
\newblock Small language model meets with reinforced vision vocabulary.
\newblock \emph{arXiv preprint arXiv:2401.12503}, 2024{\natexlab{b}}.

\bibitem[Wu et~al.(2022)Wu, Wang, Yang, Gan, Liu, Yuan, and Wang]{wu2022grit}
Jialian Wu, Jianfeng Wang, Zhengyuan Yang, Zhe Gan, Zicheng Liu, Junsong Yuan, and Lijuan Wang.
\newblock Grit: A generative region-to-text transformer for object understanding.
\newblock \emph{arXiv preprint arXiv:2212.00280}, 2022.

\bibitem[Wu et~al.(2025)Wu, Zhao, Li, Li, Zhou, Shou, and Bai]{wu2025large}
Weijia Wu, Yuzhong Zhao, Zhuang Li, Jiahong Li, Hong Zhou, Mike~Zheng Shou, and Xiang Bai.
\newblock A large cross-modal video retrieval dataset with reading comprehension.
\newblock \emph{Pattern Recognition}, 157:\penalty0 110818, 2025.

\bibitem[xAI(2024)]{grok}
xAI.
\newblock Grok, 2024.

\bibitem[Xu et~al.(2017)Xu, Zhao, Xiao, Wu, Zhang, He, and Zhuang]{mvsd}
Dejing Xu, Zhou Zhao, Jun Xiao, Fei Wu, Hanwang Zhang, Xiangnan He, and Yueting Zhuang.
\newblock Video question answering via gradually refined attention over appearance and motion.
\newblock In \emph{Proceedings of the 25th ACM international conference on Multimedia}, pp.\  1645--1653, 2017.

\bibitem[Xu et~al.(2016)Xu, Mei, Yao, and Rui]{msrvvt}
Jun Xu, Tao Mei, Ting Yao, and Yong Rui.
\newblock Msr-vtt: A large video description dataset for bridging video and language.
\newblock In \emph{Proceedings of the IEEE conference on computer vision and pattern recognition}, pp.\  5288--5296, 2016.

\bibitem[Xu et~al.(2024)Xu, Zhao, Zhou, Lin, Ng, and Feng]{pllava}
Lin Xu, Yilin Zhao, Daquan Zhou, Zhijie Lin, See~Kiong Ng, and Jiashi Feng.
\newblock Pllava: Parameter-free llava extension from images to videos for video dense captioning.
\newblock \emph{arXiv preprint arXiv:2404.16994}, 2024.

\bibitem[Yang et~al.(2024)Yang, Yang, Hui, Zheng, Yu, Zhou, Li, Li, Liu, Huang, et~al.]{yang2024qwen2}
An~Yang, Baosong Yang, Binyuan Hui, Bo~Zheng, Bowen Yu, Chang Zhou, Chengpeng Li, Chengyuan Li, Dayiheng Liu, Fei Huang, et~al.
\newblock Qwen2 technical report.
\newblock \emph{arXiv preprint arXiv:2407.10671}, 2024.

\bibitem[Yu et~al.(2024)Yu, Zhao, Wei, Yang, Wu, Kong, Wei, Wang, Ge, Zhang, et~al.]{merlin}
En~Yu, Liang Zhao, Yana Wei, Jinrong Yang, Dongming Wu, Lingyu Kong, Haoran Wei, Tiancai Wang, Zheng Ge, Xiangyu Zhang, et~al.
\newblock Merlin: Empowering multimodal llms with foresight minds.
\newblock In \emph{European Conference on Computer Vision}, pp.\  425--443. Springer, 2024.

\bibitem[Yu et~al.(2023)Yu, Yang, Li, Wang, Lin, Liu, Wang, and Wang]{mmvet}
Weihao Yu, Zhengyuan Yang, Linjie Li, Jianfeng Wang, Kevin Lin, Zicheng Liu, Xinchao Wang, and Lijuan Wang.
\newblock Mm-vet: Evaluating large multimodal models for integrated capabilities.
\newblock \emph{arXiv preprint arXiv:2308.02490}, 2023.

\bibitem[Yuan et~al.(2019)Yuan, Yu, Gu, Deng, and Li]{yuan2019novel}
Yinlong Yuan, Zhu~Liang Yu, Zhenghui Gu, Xiaoyan Deng, and Yuanqing Li.
\newblock A novel multi-step reinforcement learning method for solving reward hacking.
\newblock \emph{Applied Intelligence}, 49:\penalty0 2874--2888, 2019.

\bibitem[Yue et~al.(2024)Yue, Ni, Zhang, Zheng, Liu, Zhang, Stevens, Jiang, Ren, Sun, et~al.]{mmmu}
Xiang Yue, Yuansheng Ni, Kai Zhang, Tianyu Zheng, Ruoqi Liu, Ge~Zhang, Samuel Stevens, Dongfu Jiang, Weiming Ren, Yuxuan Sun, et~al.
\newblock Mmmu: A massive multi-discipline multimodal understanding and reasoning benchmark for expert agi.
\newblock In \emph{Proceedings of the IEEE/CVF Conference on Computer Vision and Pattern Recognition}, pp.\  9556--9567, 2024.

\bibitem[Zhai et~al.(2023)Zhai, Mustafa, Kolesnikov, and Beyer]{zhai2023sigmoid}
Xiaohua Zhai, Basil Mustafa, Alexander Kolesnikov, and Lucas Beyer.
\newblock Sigmoid loss for language image pre-training.
\newblock In \emph{Proceedings of the IEEE/CVF International Conference on Computer Vision}, pp.\  11975--11986, 2023.

\bibitem[Zhang et~al.(2023)Zhang, Li, and Bing]{video-llama}
Hang Zhang, Xin Li, and Lidong Bing.
\newblock Video-llama: An instruction-tuned audio-visual language model for video understanding.
\newblock \emph{arXiv preprint arXiv:2306.02858}, 2023.

\bibitem[Zhang et~al.(2024{\natexlab{a}})Zhang, Li, Zhang, Pu, Cahyono, Hu, Liu, Zhang, Yang, Li, et~al.]{zhang2024lmms}
Kaichen Zhang, Bo~Li, Peiyuan Zhang, Fanyi Pu, Joshua~Adrian Cahyono, Kairui Hu, Shuai Liu, Yuanhan Zhang, Jingkang Yang, Chunyuan Li, et~al.
\newblock Lmms-eval: Reality check on the evaluation of large multimodal models.
\newblock \emph{arXiv preprint arXiv:2407.12772}, 2024{\natexlab{a}}.

\bibitem[Zhang et~al.(2022)Zhang, Sun, Jiang, Yu, Weng, Yuan, Luo, Liu, and Wang]{zhang2022bytetrack}
Yifu Zhang, Peize Sun, Yi~Jiang, Dongdong Yu, Fucheng Weng, Zehuan Yuan, Ping Luo, Wenyu Liu, and Xinggang Wang.
\newblock Bytetrack: Multi-object tracking by associating every detection box.
\newblock In \emph{European conference on computer vision}, pp.\  1--21. Springer, 2022.

\bibitem[Zhang et~al.(2024{\natexlab{b}})Zhang, Tzeng, Du, and Kislyuk]{zhang2024large}
Yinan Zhang, Eric Tzeng, Yilun Du, and Dmitry Kislyuk.
\newblock Large-scale reinforcement learning for diffusion models.
\newblock \emph{arXiv preprint arXiv:2401.12244}, 4, 2024{\natexlab{b}}.

\bibitem[Zhang et~al.(2024{\natexlab{c}})Zhang, Li, Liu, Lee, Gui, Fu, Feng, Liu, and Li]{llavanext-video}
Yuanhan Zhang, Bo~Li, haotian Liu, Yong~jae Lee, Liangke Gui, Di~Fu, Jiashi Feng, Ziwei Liu, and Chunyuan Li.
\newblock Llava-next: A strong zero-shot video understanding model, April 2024{\natexlab{c}}.
\newblock URL \url{https://llava-vl.github.io/blog/2024-04-30-llava-next-video/}.

\bibitem[Zhao et~al.(2023{\natexlab{a}})Zhao, Cai, Si, Ma, An, Chen, Liu, Wang, Han, and Chang]{zhao2023mmicl}
Haozhe Zhao, Zefan Cai, Shuzheng Si, Xiaojian Ma, Kaikai An, Liang Chen, Zixuan Liu, Sheng Wang, Wenjuan Han, and Baobao Chang.
\newblock Mmicl: Empowering vision-language model with multi-modal in-context learning.
\newblock \emph{arXiv preprint arXiv:2309.07915}, 2023{\natexlab{a}}.

\bibitem[Zhao et~al.(2023{\natexlab{b}})Zhao, Yu, Ge, Yang, Wei, Zhou, Sun, Peng, Dong, Han, et~al.]{chatspot}
Liang Zhao, En~Yu, Zheng Ge, Jinrong Yang, Haoran Wei, Hongyu Zhou, Jianjian Sun, Yuang Peng, Runpei Dong, Chunrui Han, et~al.
\newblock Chatspot: Bootstrapping multimodal llms via precise referring instruction tuning.
\newblock \emph{arXiv preprint arXiv:2307.09474}, 2023{\natexlab{b}}.

\bibitem[Zhou et~al.(2018)Zhou, Xu, and Corso]{zhou2018towards}
Luowei Zhou, Chenliang Xu, and Jason Corso.
\newblock Towards automatic learning of procedures from web instructional videos.
\newblock In \emph{Proceedings of the AAAI Conference on Artificial Intelligence}, volume~32, 2018.

\bibitem[Zhu et~al.(2023)Zhu, Chen, Shen, Li, and Elhoseiny]{minigpt4}
Deyao Zhu, Jun Chen, Xiaoqian Shen, Xiang Li, and Mohamed Elhoseiny.
\newblock Minigpt-4: Enhancing vision-language understanding with advanced large language models, 2023.

\bibitem[Zhu et~al.(2025)Zhu, Zhao, Lin, Yang, Yu, Liu, Wei, Sun, Ge, and Zhang]{zhu2025perpo}
Zining Zhu, Liang Zhao, Kangheng Lin, Jinze Yang, En~Yu, Chenglong Liu, Haoran Wei, Jianjian Sun, Zheng Ge, and Xiangyu Zhang.
\newblock Perpo: Perceptual preference optimization via discriminative rewarding.
\newblock \emph{arXiv preprint arXiv:2502.04371}, 2025.

\end{thebibliography}
\bibliographystyle{iclr2025_conference}

\newpage
\appendix
\section{Appendix}
\label{appendix}

In this appendix, we provide additional details about \textit{temporal hacking} and our \textit{\textbf{U}nhackable \textbf{T}emporal \textbf{R}ewarding (\textbf{UTR})}, which were omitted due to the 10-page limit of the main paper. Specifically, Section~\ref{exp_detail} elaborates on the dataset and training settings of Video-UTR. Section~\ref{add_exp} presents additional experiments to analyze UTR’s characteristics. Section~\ref{more_cases} offers more qualitative examples to demonstrate the capabilities of Video-UTR, and Section~\ref{add_disscusion} provides further discussion of existing approaches.

\section{Additional Details about Experimental Setting}
\label{exp_detail}

\textbf{Additional information of the datasets.} In Section~\ref{UTR} of the manuscript, we introduced how we established the unhackable temporal rewarding (UTR) including \textit{data modeling (UTR-Data) and \textit{task modeling (Bidirectional Querying)}}. Now, in this section, we go into greater detail about how we collected and built the UTR-Data and how we constructed task conversation. To start, we provide an overview of our collected data in Table~\ref{tab:train_data}, and then dive into the step-by-step process of how it was constructed.

\begin{table}[h]
\centering                         
\renewcommand{\arraystretch}{1.4}  
\caption{\textbf{Training Data Statistics.} We first build our UTR-Data mainly based on sampled HowTo100M, MeViS, and LaMOT. Then we mix UTR-Data with several existing video conversation data, \ie, LLaVA-NeXT-SFT and VideoChat2.}
\setlength{\tabcolsep}{1.5mm}      
  \resizebox{1.0\columnwidth}{!}{
\small                             
\begin{tabular}{c | c |c c c | c}
\toprule
\textbf{Modality} & \multicolumn{1}{c|}{\textbf{Dataset}} & \textbf{Original} & \textbf{Used} & \textbf{Ratio\%} & Training Stage \\ 
\hline
 & HowTo100M~\citep{miech2019howto100m} & 100M & 50K & 0.05\% & \text{Stage II} \\
 & MeViS~\citep{ding2023mevis}       & 443K & 90K & 20.3\% & \text{Stage II}\\
 & LaMOT~\citep{li2024lamot} & 2.44M  & 225K & 10.5\% & Stage II \\ 
\multirow{-4}{*}{Video-Text} & VideoChat2~\citep{mvbench} & 2M & 100K & 5\% & Stage II \\ 
\midrule
 & BLIP-558K~\citep{llava} & 558K & 558K & 100\% & Stage I\\
\multirow{-2}{*}{Image-Text} & LLaVA-NeXT-SFT~\citep{llavanext-video} & 790K & 790K & 100\% & Stage II                     \\ 
\hline
Vision-Language & Total & 106.231M & 1.813M & 1.71\% & Stage I \& II  \\
\bottomrule
\end{tabular}
\vspace{3mm}
}
\label{tab:train_data}
\end{table}

Specifically, we follow the steps below to pre-process the raw video data to construct UTR-Data

\noindent (1) Randomly sample the fixed number ($16$, $24$ or $32$) frames at a certain frame (gap = 3,4 or 5) or random interval to form a video clip each time.

\noindent (2) Extract all spatiotemporal attribution trajectories containing their category, identity, action and bounding boxes in each video clip. This can be accomplished through expert models, \eg, GRiT~\citep{wu2022grit}, Grounding DINO~\citep{liu2023grounding}, and ByteTrack~\citep{zhang2022bytetrack} or directly obtained from the annotations provided by datasets.

\noindent (3) Remove the trajectory containing too small objects (smaller than $1/32$ of the image size).

\noindent (4) Random select observation (spatial or temporal attributions in the randomly selected frame) as the query to conduct bidirectional querying task modeling.

\noindent (5) Compose the task format as the following: 

\texttt{\textit{\textbf{Question}: System prompt + query question}}. 

\texttt{\textit{\textbf{Answer}: query answer, cat1<idi>Frame1:<box>;Frame2:<box>;...</idi>}},

where \texttt{<query question, query answer>} is the question-answer pair that is designed based on the selected querying attributes. 

\textbf{Additional Training Setting Details.} As stated in the manuscript, Video-UTR follows a two-stage training procedure. In this part, we will provide a detailed overview of our training settings, including the hardware used for training, the duration, and the training hyperparameters. All information are recoderd in Table~\ref{tab:hyperparam}.

\begin{table}[t]
    \centering
    \caption{\textbf{Training hyperparameters of Video-UTR}. The hyperparameter placed in the middle indicates that this hyperparameter is used in both stages.}
    \begin{tabular}{l cc}
         \toprule
         \textbf{Configuration}            & \textbf{Stage I} & \textbf{Stage II} \\
         \midrule
         Machine                  & \multicolumn{2}{c}{NVIDIA Tesla A800 80GB GPU x 64}\\
         Training hours           & 1 hour   & 20 hours\\
         \midrule
         ViT init.                & SigLIP-so400m-patch14-384 & Video-UTR Stage I\\
         LLM init.                & Qwen2-7B-Instruct & Video-UTR Stage I \\
         Projection init.         & random & Video-UTR Stage I \\
         Image resolution         & $384^2$ & $384^2$ \\
         ViT sequence length      & 2048 & 2048 \\
         LLM sequence length      & 32K & 32K\\
         Video Frame length       & 1 & 32 \\
         Optimizer                & \multicolumn{2}{c}{AdamW} \\
         Optimizer hyperparameter & \multicolumn{2}{c}{$\beta_{2}=0.95, eps=1e^{-8}$} \\
         Peak learning rate       & \multicolumn{2}{c}{Vision Tower:  $2e^{-6}$;  LLM: $1e^{-5}$} \\
         Minimum learning rate    & \multicolumn{2}{c}{0} \\
         ViT learning rate decay  & 0.9 & 0 \\
         ViT Drop path rate       & \multicolumn{2}{c}{0} \\
         Learning rate schedule   & \multicolumn{2}{c}{cosine decay} \\
         Weight decay             & \multicolumn{2}{c}{0.05} \\
         Gradient clip            & \multicolumn{2}{c}{1.0} \\
         Training steps           & 1k & 5k \\
         Warm-up ratio            & 0.003 & 0.003 \\
         Global batch size        & 512 & 256 \\
         Gradient Acc.            & 1 & 4 \\
         Numerical precision      & \multicolumn{2}{c}{$\mathtt{bfloat16}$} \\
         Optimizer sharding       & \multicolumn{2}{c}{\ding{51}} \\
         Activation checkpointing & \multicolumn{2}{c}{\ding{55}} \\
         Model parallelism        & \multicolumn{2}{c}{\ding{55}} \\
         Pipeline parallelism     & \multicolumn{2}{c}{\ding{55}} \\
         \bottomrule
    \end{tabular}
    \label{tab:hyperparam}
\end{table}

\textbf{Additional Testing Setting Details.} In the inference and evaluation phase, we essentially follow the settings of PLLaVA~\citep{pllava} and LLaVA-NeXT-Video~\citep{llavanext-video}, including the system prompt for inference, the number of frames extracted, and so on, while conducting evaluations on the existing video benchmark. Specifically, as illustrated in Table~\ref{tab:text_setting}, we mainly use the uniform frame sampling for frame selection. For answer selection based on GPT score, we mainly use the gpt-3.5-turbo-0613 version to evaluate the responses of our model.

\begin{table}[t]
\centering                         
\renewcommand{\arraystretch}{1.4}  
\caption{\textbf{Video benchmark evaluation setting.} We report some detailed setting during evaluation. MCQ: Multi-choice question. QA: Question-answer.}
\setlength{\tabcolsep}{1.5mm}      
  \resizebox{1.0\columnwidth}{!}{
\large                          
\begin{tabular}{c | c |c |c | c}
\toprule
\textbf{Benchmark} & \multicolumn{1}{c|}{\textbf{Evaluation type}} & \textbf{Prompts} & \textbf{Input frames} & \textbf{Answer selection} \\ 
\hline
 & MCQ & \textbf{\texttt{Question}} + ``Please directly give the best option:" & 32 & \\
 & Yes or No       &  \textbf{\texttt{Question}} + ``Please answer yes or no:" & 32 & \\
 & Caption Matching &  \textbf{\texttt{Question}} + ``Please directly give the best option:"  & 32 &  \\ 
\multirow{-4}{*}{\makecell{Temoral\\Compass}} & Captioning &  \textbf{\texttt{Question}} & 32 & \multirow{-4}{*}{GPT score} \\ 
\midrule
MVBench & Video - MCQ & "Question" +  \textbf{\texttt{Question}} +``Option:" + \textbf{\texttt{Options}} + ``Only give the best option." & 32 & Option matching                     \\ 
\hline
& & & & \\
VideoMME & Video-MCQ &   \multirow{-2}{*}{\makecell{These are the frames of a video. Select the best answer to the following multiple-choice \\ question based on the video. Respond with only the letter (A, B, C, or D) of the correct option.}} & 32 & Option matching  \\
\hline
MSVD & Video QA&   \textbf{\texttt{Question}}& 32& GPT Score\\
\hline
ActivityNetQA & Video QA &    \textbf{\texttt{Question}}+ ``Answer the question using a single word or phrase." & 32& GPT Score\\
\hline
TGIFQA & Video QA & \textbf{\texttt{Question}}& 32& GPT Score\\
\hline
VideoChatGPT & Video QA &  \textbf{\texttt{Question}}& 32& GPT Score\\
\bottomrule
\end{tabular}
}
\label{tab:text_setting}
\end{table}

\section{Additional Experimental Analysis}
\label{add_exp}

\begin{wrapfigure}{r}{9.5cm} 
    \vspace{-0.2cm} 
    \centering
    \footnotesize
    \includegraphics[width=0.67\columnwidth]{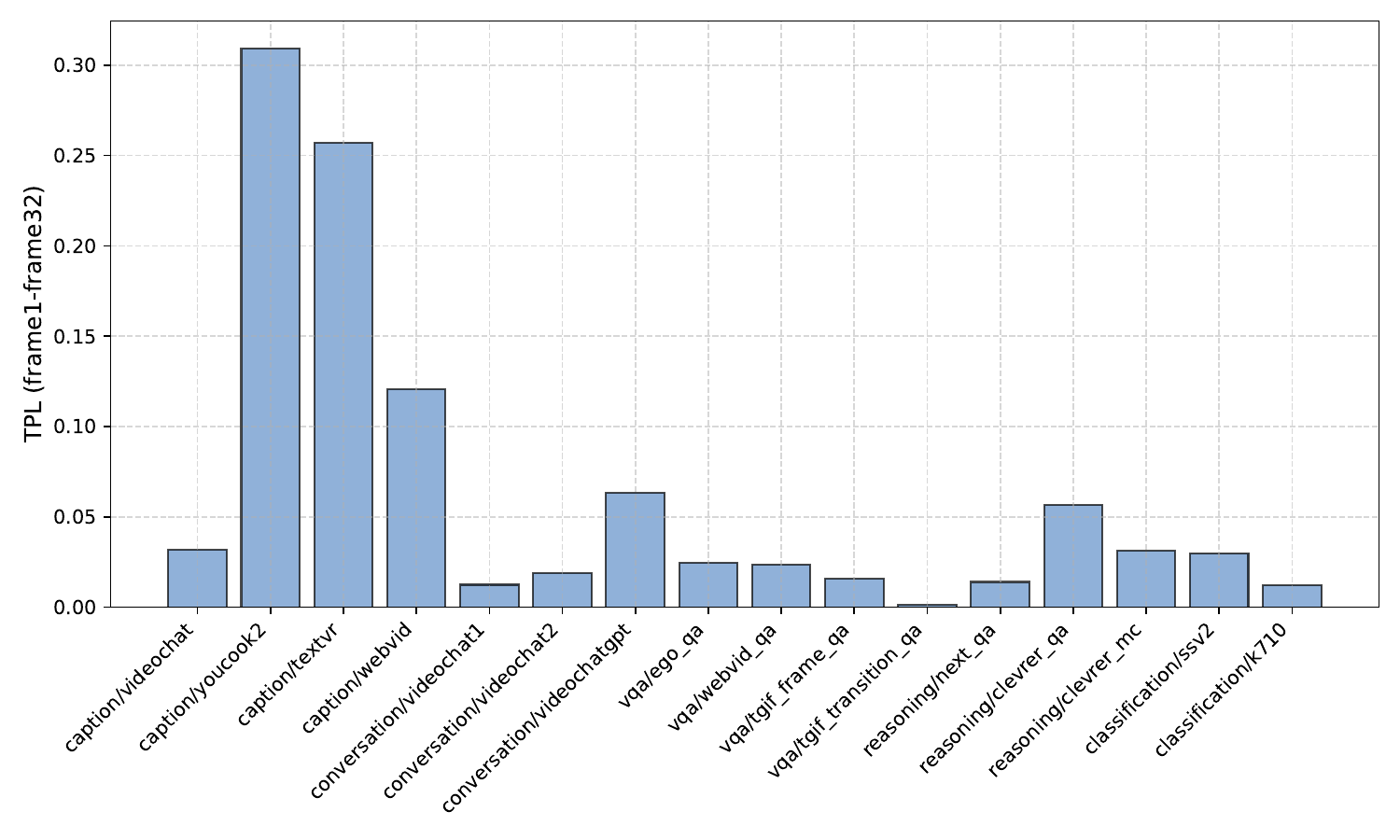}
    \vspace{-5mm} 
    {
        \caption{\textbf{Quantitative TPL statistic} of VideoChat2. \vspace{-0.2cm} }
        \label{fig: tp_videochat2}
    }
\end{wrapfigure}

\textbf{More analysis about temporal perplexity (TPL).} In the Section~\ref{exp_tp}, we present a case study to illustrate the relationship between the proposed TPL score and data quality, where a higher TPL score indicates better data quality. In this part, we further present the relationship between TPL and data quality from a quantitative statistical perspective. Specifically, we calculate the TPL score for different data subsets in VideoChat2~\citep{mvbench} and computed their average values. The results are shown in Figure~\ref{fig: tp_videochat2}. We can observe that the TPL distribution for the YouCook2~\citep{zhou2018towards} and TextVR~\citep{wu2025large} subset is relatively high. This suggests that these two data subsets are of relatively high quality. As we know, these datasets, such as YouCook2, contain a large amount of first-person perspective and high-motion video data. These videos are rich in high information density and dynamic content, which is beneficial for the model’s temporal modeling. The results further prove that TPL provides a reference for selecting high-quality data from VideoChat2. Based on the TPL distribution, sampling more reasoning data is likely to be more beneficial for achieving better video-language modeling.

\begin{figure*}[t!]
\centering
\includegraphics[width=1.0\linewidth]{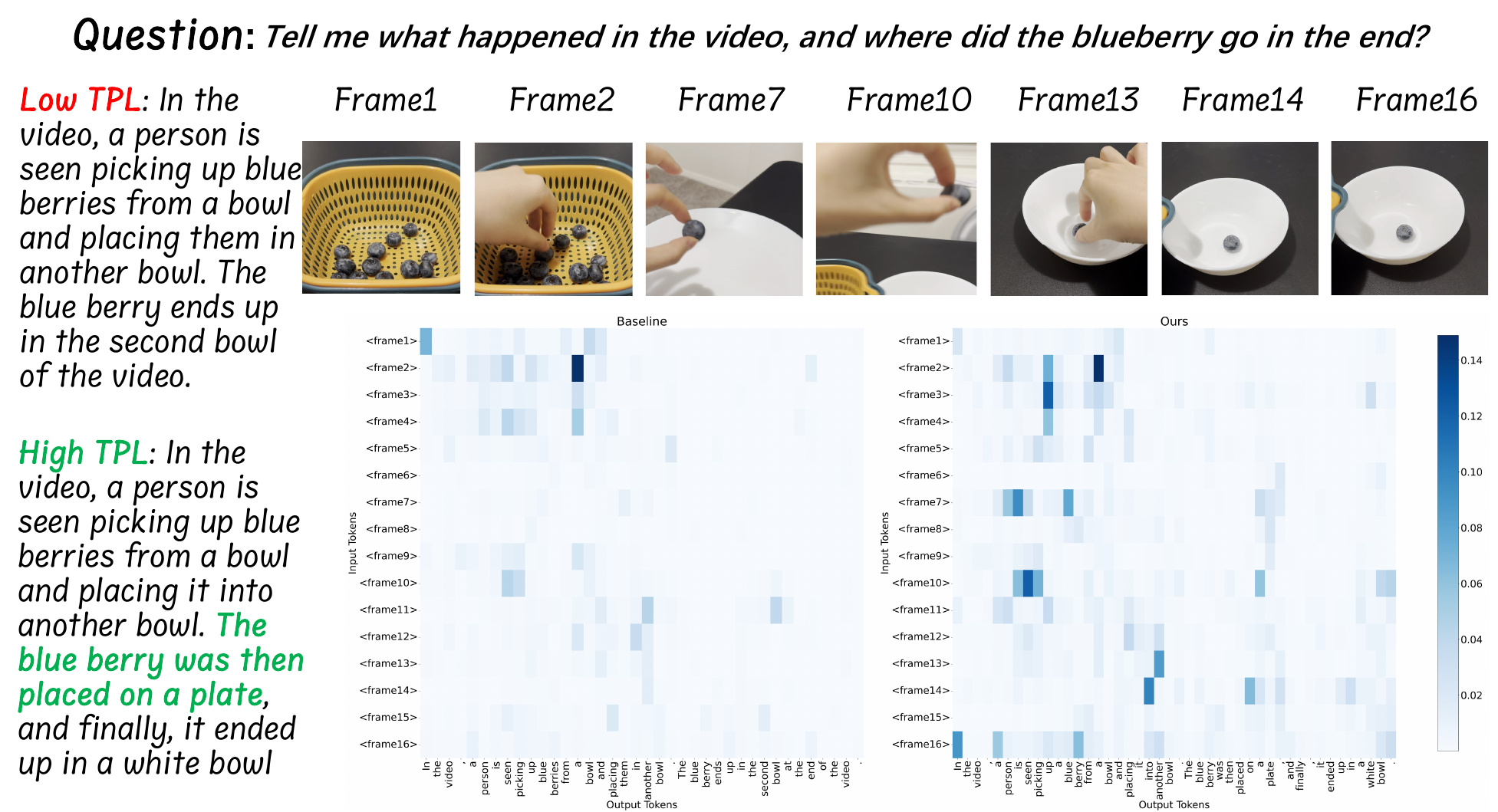}
\caption{\textbf{Output attention visualization.} We compute the average output layer attention of the tokens generated by the model for each frame in the QA task and visualized the results.}
\label{fig:attn_plus1}
\vspace{-4mm}
\end{figure*}

\begin{figure*}[t!]
\centering
\includegraphics[width=1.0\linewidth]{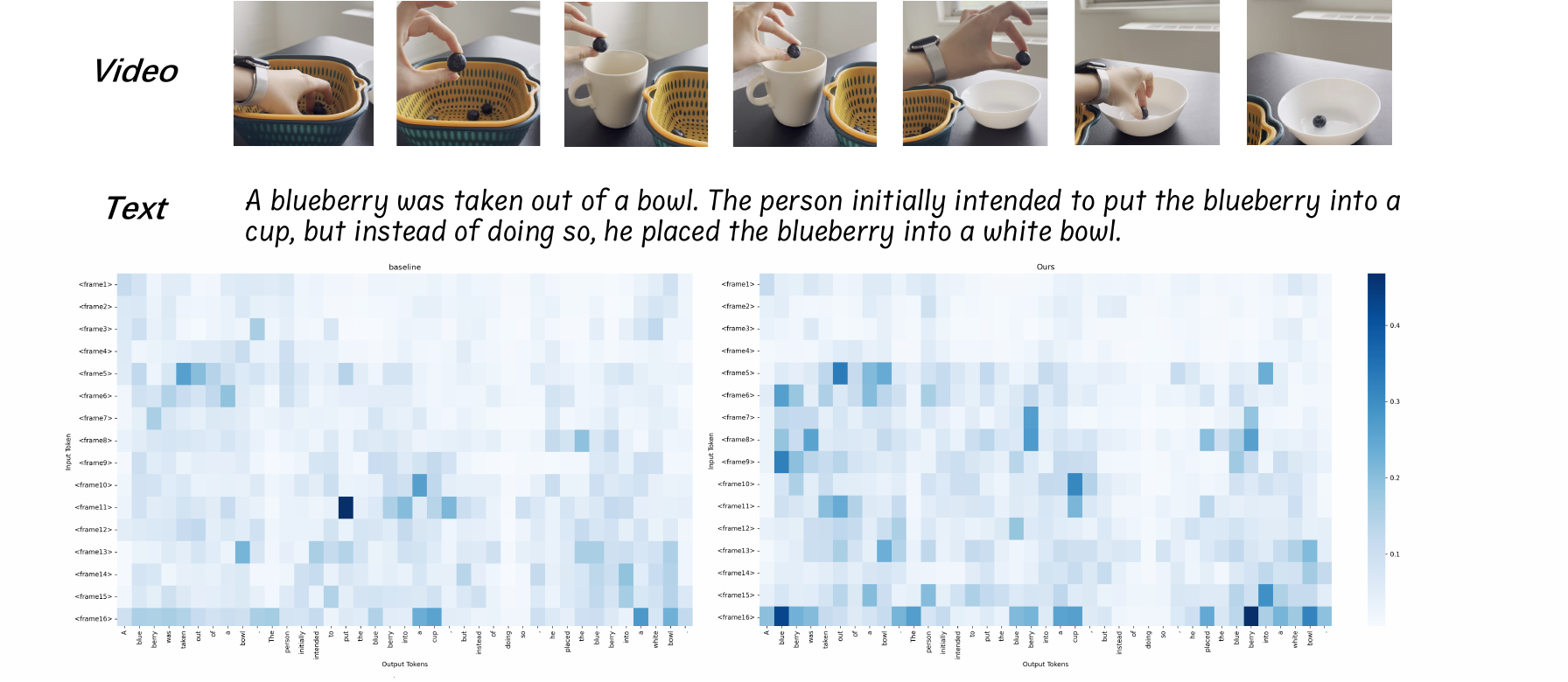}
\caption{\textbf{Video-text input attention visualization.} The left is the attention map of the model with low TPL while the right is the attention map with high TPL score.}
\label{fig:attn_plus2}
\vspace{-4mm}
\end{figure*}

\textbf{More attention visualization analysis.} In Figure~\ref{fig:attnmap}, we present the attention map visualizations of frame tokens under model responses at different TPL levels. In this part, we further provide more detailed attention analysis. As shown in Figure~\ref{fig:attn_plus1} \&~\ref{fig:attn_plus2}, we conduct two forms of attention visualization. The first involves video QA, visualizing the attention values between the answer content and the tokens of each video frame. The second form calculates the self-attention when inputting the video-text pair into the model simultaneously. From both visualization results, we can observe that our Video-UTR, while achieving a higher TPL score, clearly attends to more frames, thereby avoiding the loss of crucial details in the video and making the answers more accurate and detailed.

\begin{figure*}[t!]
\centering
\includegraphics[width=1.0\linewidth]{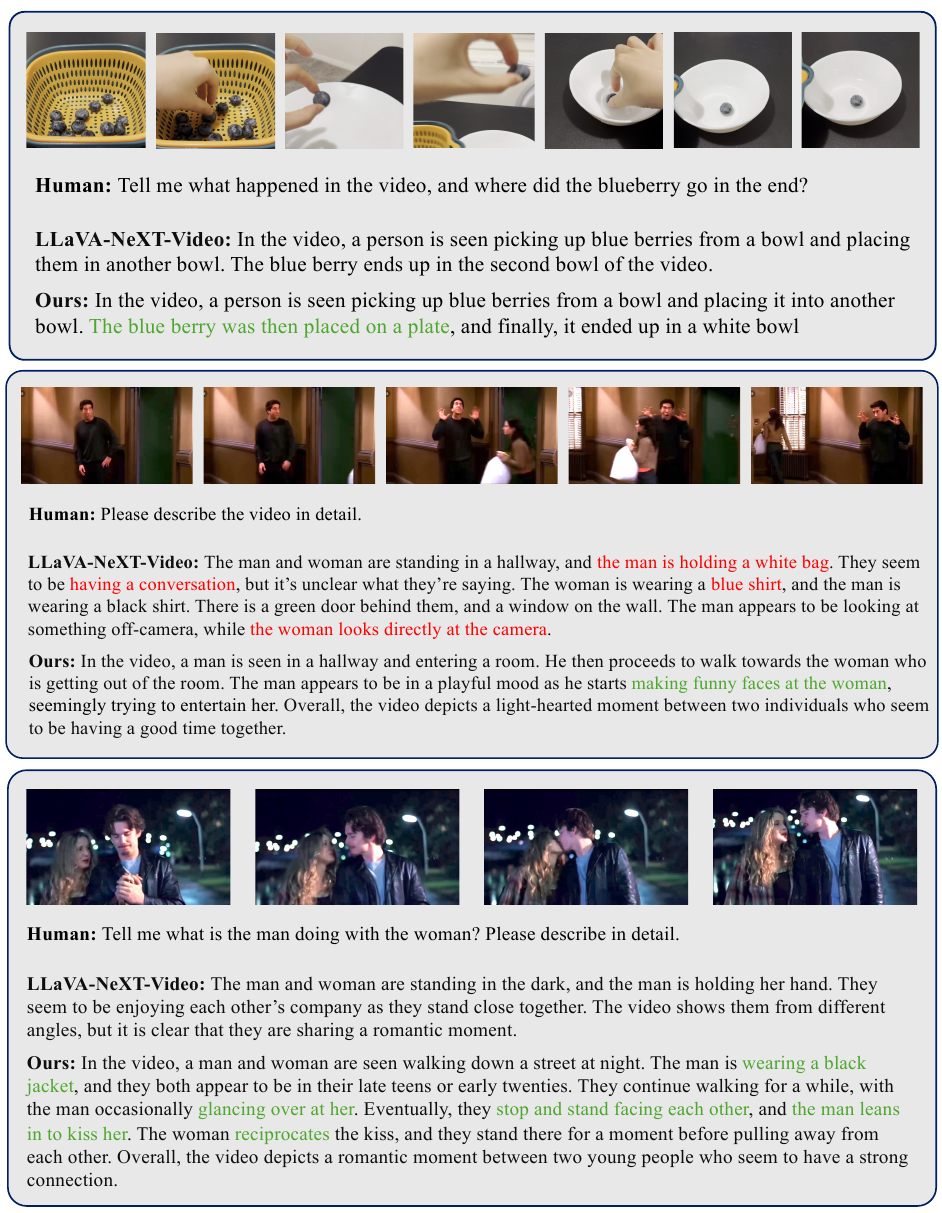}
\caption{\textbf{Qualitative examples visualization} of Video-UTR. Please note that we only display the most important frames from the full video (32 frames) to conserve space.}
\label{fig:demo_case}
\end{figure*}

\section{Additional Qualitative Analysis}
\label{more_cases}

In this section, to more intuitively demonstrate the unhackable capability of Video-UTR, we present several subjective video Q\&A cases, as shown in Figure~\ref{fig:demo_case}. Compared to our baseline, LLaVA-Next-Video, our Video-UTR demonstrates a more accurate video understanding capability, specifically by better comprehending user queries, focusing on more video details, and providing more precise and less hallucinated responses. These results further validate the effectiveness of our proposed UTR modeling.

\section{More Disscusions}
\label{add_disscusion}

\subsection{Dependency on Expert Models for UTR}
\label{add_utr}

As introduced in Section~\ref{UTR} of the main text, UTR leverages existing expert models to extract spatiotemporal attribute cues, which serve as the foundation for data modeling and task modeling. Therefore, the capability of the selected expert models and the quality of the extracted attributes are critical variables that significantly influence the effectiveness of UTR modeling. In this part, we will delve into the significance of selecting expert models, the selection criteria, the details of attribute extraction, and the validation of attribute quality.

\textbf{The importance and rationale behind selecting expert models for attribute extraction.} The use of expert models to support MLLM training has become a widely adopted strategy in the current development stage. Notable implementations include models such as PaLI-X~\citep{chen2023pali}, Qwen-VL~\citep{qwen}, InternVL~\citep{chen2024internvl}, and LLaMA3.2-Vision~\citep{llama3.2}, which integrate domain-specific expert models spanning areas like detection, grounding, and OCR to scale up training data annotation. The effectiveness of this approach has been well-validated through extensive empirical studies. Fundamentally, these pipelines operate as a distillation process, transferring knowledge from expert models to MLLMs to enhance specific capabilities, such as fine-grained perception. In line with this paradigm, our proposed UTR framework employs expert models to extract spatiotemporal attributes from video data, thereby strengthening the spatiotemporal perception abilities of video MLLMs. This improvement is substantiated by the empirical results presented in Table~\ref{exp_st} of our manuscript.

\textbf{Extraction and filtering of high-quality attributes.} To select specific expert models, we conducted a systematic evaluation based on existing benchmarks, \eg, COCO~\citep{coco}, Lvis~\citep{gupta2019lvis}, VG~\citep{krishna2017visual}, \etc., of the performance of various options, such as GRiT~\citep{wu2022grit} and GroundingDINO~\citep{liu2023grounding}, to identify the most suitable candidates. For the proposed spatiotemporal attributes—including bounding boxes, captions, identities, and actions, as illustrated in Figure~\ref{fig:UTR} of our manuscript—we implemented a multi-stage selection and filtering process. \textbf{First}, we filtered the attributes based on the \textit{confidence scores} provided by the expert models. \textbf{Next}, we applied a multi-object tracking algorithm, \ie, ByteTrack~\citep{zhang2022bytetrack}, to analyze contextual correlations within the video content. This analysis included examining factors such as the Intersection over Union (IoU) of bounding boxes across frames and trajectory continuity metrics, ensuring that trajectory lengths exceeded predefined thresholds. This comprehensive process ensures the reliability and consistency of the extracted attribute trajectories, thereby enhancing their overall quality and utility.

\begin{wraptable}{r}{0.57\textwidth}
    \caption{\textbf{Human validation} of extracted attributes.}
    \vspace{-1em}
    \label{tab:human_quality}
    \begin{center}        
    \def\arraystretch{1}
    \resizebox{\linewidth}{!}{
    \begin{tabular}{c|ccc}
        \toprule
        \bf Validation & \bf Location & \bf Description & \bf Consistency \\
        \hline
        Human & 2.98 & 2.23 & 2.57\\
        \bottomrule
    \end{tabular}
    \vspace{-20em}
    }
    \end{center}
\end{wraptable}

\textbf{Human validation of the extracted attributes.} To further validate the effectiveness of the extracted spatiotemporal attributes from video data, we conducted a human evaluation experiment. Specifically, 100 data samples generated using our UTR pipeline were randomly selected for assessment by human evaluators.
Human annotators will score these data based on three criteria: the accuracy of the subject bounding box, the correctness of the attribute descriptions, and the consistency of the attribute trajectories, using a scoring range of 1 to 3. The results is shown in Table~\ref{tab:human_quality}. We can observe that the average quality score of the extracted attributes is quite high, indicating a strong level of reliability. The results of this evaluation highlight the robustness and high quality of both the extracted spatiotemporal attributes and the constructed data, confirming the reliability of our pipeline.

\subsection{Consistency of TPL with Human Judgment}
\label{add_tpl}

\begin{table}[h]
    \caption{\textbf{Consistency between TPL score and human judgment}.}
    \vspace{-1em}
    \label{tab:tpl_consis}
    \begin{center}        
    \resizebox{0.9\linewidth}{!}{
    \begin{tabular}{c|ccc ccc cc}
        \toprule
          \multirow{2}{*}{\textbf{Validation}} & \multicolumn{2}{c}{High} &&\multicolumn{2}{c}{Medium} &&\multicolumn{2}{c}{Low}\\
    \cmidrule{2-3}  \cmidrule{5-6}  \cmidrule{8-9}
    
        & \bf Richness & \bf Relevance  && \bf Richness & \bf Relevance &&  \bf Richness & \bf Relevance\\
        \hline
        TPL level & 3 & 3 && 2 & 2 && 1 & 1\\
        Human & 2.85 & 2.76 && 2.15 & 1.85 && 1.61 & 1.64\\
        \bottomrule
    \end{tabular}
    \vspace{-10em}
    }
    \end{center}
        \vspace{-2em}
\end{table}

In Section~\ref{exp_tp}, we point out that TPL not only reflects the degree of temporal hacking in the video-language modeling process, but it can also serve as a high-order metric to indicate the quality of video-text pairs. In this part, we plan to further explore this issue by examining the consistency of TPL with human judgment, highlighting the reliability of TPL score as a data filtering metric.

Specifically, we first randomly select 100 video-text pairs from VideoChat2~\citep{li2023videochat} and calculate their temporal perplexity based on the definition in Eq.~\ref{eq:tp}. Next, we sort the data by their TPL values and divide it into three groups: high, medium, and low. We then invite several human annotators to rate these sampled video-text pairs on a scale of 1 to 3. The criteria for scoring includes two aspects, \ie, the richness of the video-text information (considering both information density and dynamics) and the relevance of the video to the text. Based on the annotators’ scores, the consistency can be evaluated based on the average human ratings and their alignment with the level categories. Before the human annotators begin the annotation process, we provide a detailed annotation guideline, which explains the scoring criteria and standards comprehensively and includes relevant references. Specifically, the two composite metrics, \ie, richness and relevance, are defined as follows:

\begin{itemize}[leftmargin=2.5mm]
\setlength{\itemsep}{2pt}

    \item \textbf{Richness:}
        \begin{itemize}[leftmargin=2.5mm]
        \setlength{\itemsep}{2pt}
            \item Frame information density, referring to the degree to which each frame corresponds to an independent description.
            \item Level of descriptive detail, referring to the richness of details included from the video.
            \item The richness of motion information, including the extent of motion in both the subject and the scene.
        \end{itemize}
    
    \item \textbf{Relevance:}
        \begin{itemize}[leftmargin=2.5mm]
        \setlength{\itemsep}{2pt}
            \item The relevance between the video and the text, specifically, the degree to which the description corresponds to the video content.
            \item The relevance of the video context, specifically, the extent to which the descriptions of relevant subjects in the video reflect dynamic changes.
        \end{itemize}
    
\end{itemize}

Follow the above standards, annotators conducted the validation of the sampled cases and ultimately summarized their assessments into scores for two composite metrics. The results is shown in Table~\ref{tab:tpl_consis}. We can observe that the groupings based on TPL scores and those based on human judgments are generally consistent. This indicates that our proposed TPL score is a reliable metric for filtering high-quality video-text pair data.

\subsection{Failure Case Analysis of UTR}
\label{failure_case}

\begin{figure*}[t!]
\centering
\includegraphics[width=1.0\linewidth]{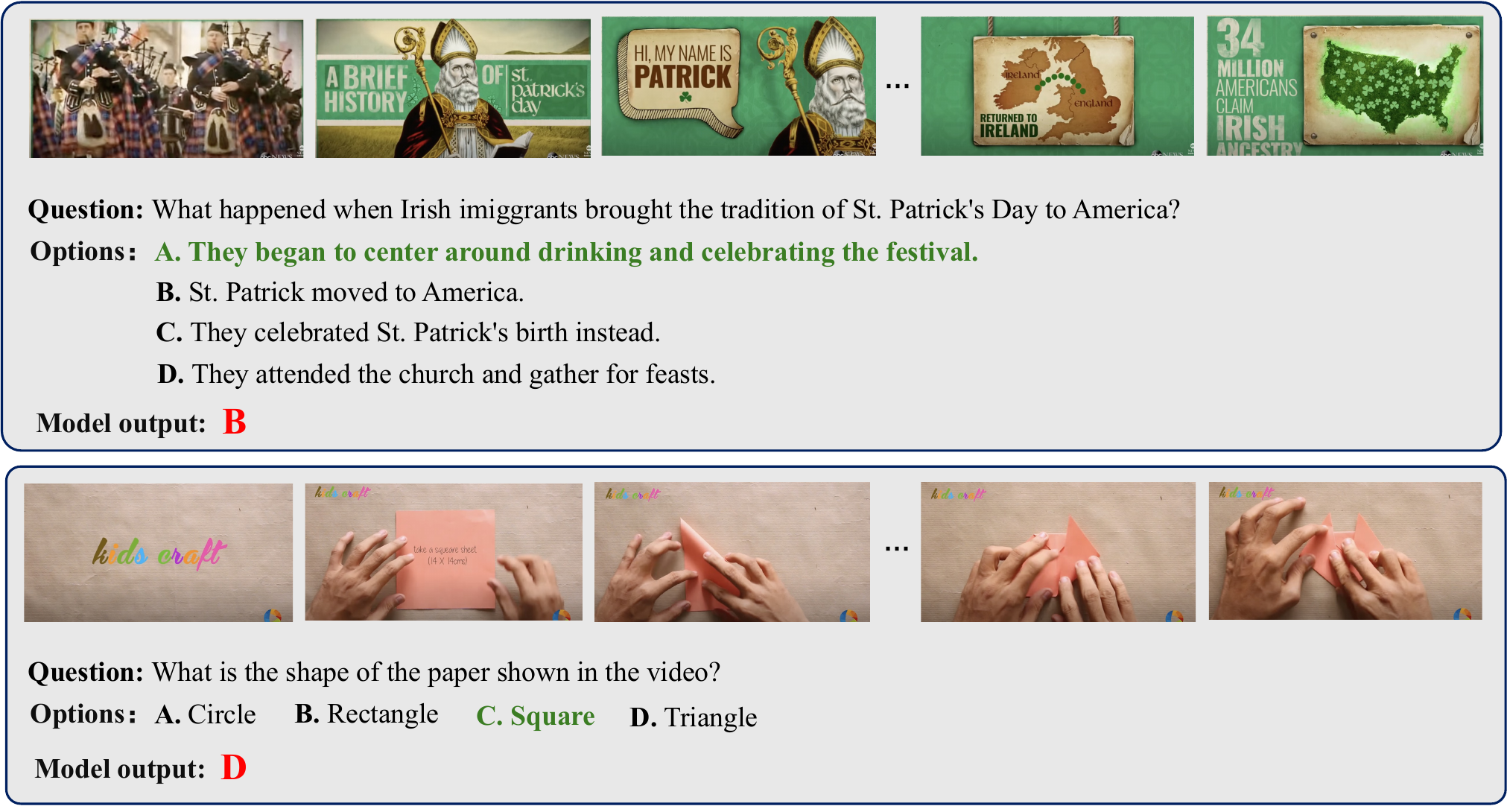}
\caption{\textbf{Failure case visualization} of Video-UTR. We select two representative failure cases from the VideoMME~\citep{videomme} benchmark.}
\label{fig:failure_case}
\vspace{-4mm}
\end{figure*}

Although our proposed UTR significantly mitigates temporal hacking from both data modeling and task modeling perspectives, it still has limitations in some situation, and we identify several representative examples on the VideoMME~\citep{videomme} benchmark. As illustrated in Figure~\ref{fig:failure_case}, the top case shows that Video-UTR does not perform as well on certain knowledge-oriented Video MCQ tasks. This type of question tests the inherent knowledge base of large language models, so our UTR method does not result in a significant improvement. The bottom case illustrates that in scenarios where the answer can be determined by analyzing a single frame or a few frames, our UTR method does not demonstrate a significant advantage. Placing more emphasis on the overall video content does not provide notable benefits in addressing such questions.

The aforementioned failure cases analysis also highlights the need to design better video understanding benchmarks that can more reasonably and reliably evaluate the ability of video MLLMs to observe and comprehend the overall video content, rather than relying heavily on the inherent capabilities of LLMs.

\subsection{Limitaion and Furture Work.}

\textbf{Limitation of Unhackable Temporal Hacking.} Although our proposed UTR significantly mitigates temporal hacking from both data modeling and task modeling perspectives, it has a noticeable limitation in terms of its reliance on expert model accuracy. Since UTR modeling is based on extracted subject attributes, the quality of these attributes—such as positional accuracy, precise descriptions of the subject’s appearance and actions, and the accuracy of trajectory associations—directly impacts the overall performance of the final model. Therefore, improving the quality of these extracted subject attributes represents a highly valuable direction for future improvement. 

\textbf{Future work.} On the other hand, seamlessly integrating the constructed attribute trajectories into dialogues poses yet another challenging issue. Exploring whether a single multimodal large language model~\citep{merlin, dreamllm, peng2024dreambench++} can be utilized to handle the entire data processing and task construction pipeline is a highly promising research direction. In addition, the establishment of a more explicit temporal rewarding mechanism represents a valuable research direction. Leveraging reinforcement learning algorithms, \eg, DPO~\citep{rafailov2024direct, zhu2025perpo}, PPO~\citep{schulman2017proximal} and GRPO~\citep{r1}, for post-training enhancement of the model's video comprehension and reasoning capabilities constitutes a key focus for future research.

\end{document}